\documentclass[sigconf]{acmart}

\AtBeginDocument{%
  }

\usepackage{multirow}
\usepackage[ruled,vlined,linesnumbered]{algorithm2e}
\usepackage{tikz}
\usepackage{
amsmath}
\usepackage{float}
\usepackage{filecontents}
\usepackage
{booktabs}
\usepackage{graphicx}
\usepackage{subcaption}
\usepackage{amsfonts}
\usepackage{hyperref}
\usepackage{enumitem}

\usepackage{color}

\setcopyright{acmlicensed}
\copyrightyear{2018}
\acmYear{2018}
\acmDOI{XXXXXXX.XXXXXXX}

\acmConference[Conference acronym 'XX]{Make sure to enter the correct
  conference title from your rights confirmation emai}{June 03--05,
  2018}{Woodstock, NY}
\acmISBN{978-1-4503-XXXX-X/18/06}

\begin{document}

\title{On the Robustness of Graph Reduction Against GNN Backdoor}


\author{Yuxuan Zhu}
\author{Michael Mandulak}
\author{Kerui Wu}
\affiliation{ 
      \institution{Rensselaer Polytechnic Institute}
      \city{Troy, NY}
      \country{USA}
    }
\email{{zhuy27,mandum, wuk9}@rpi.edu}


\author{George Slota} 
\affiliation{ 
      \institution{Rensselaer Polytechnic Institute}
      \city{Troy, NY}
      \country{USA}
    }
\email{slotag@rpi.edu}

\author{Yuseok Jeon} 
\affiliation{ 
      \institution{Ulsan National Institute of Science and Technology}
      \city{Ulsan}
      \country{South Korea}
    }
\email{ysjeon@unist.ac.kr}

\author{Ka-Ho Chow} 
\affiliation{ 
      \institution{The University of Hong Kong}
      \city{Hong Kong}
      \country{China}
    }
\email{kachow@cs.hku.hk}

\author{Lei Yu} 
\affiliation{ 
      \institution{Rensselaer Polytechnic Institute}
      \city{Troy, NY}
      \country{USA}
    }
\email{yul9@rpi.edu}


\begin{abstract}
Graph Neural Networks (GNNs) are gaining popularity across various domains due to their effectiveness in learning graph-structured data. Nevertheless, they have been shown to be susceptible to backdoor poisoning attacks, which pose serious threats to real-world applications. Meanwhile, graph reduction techniques, including coarsening and sparsification, which have long been employed to improve the scalability of large graph computational tasks, have recently emerged as effective methods for accelerating GNN training on large-scale graphs. However, the current development and deployment of graph reduction techniques for large graphs overlook the potential risks of data poisoning attacks against GNNs. It is not yet clear how graph reduction interacts with existing backdoor attacks.
This paper conducts a thorough examination of the robustness of graph reduction methods in scalable GNN training in the presence of state-of-the-art backdoor attacks. We performed a comprehensive robustness analysis across six coarsening methods and six sparsification methods for graph reduction, under three GNN backdoor attacks against three GNN architectures. Our findings indicate that the effectiveness of graph reduction methods in mitigating attack success rates varies significantly, with some methods even exacerbating the attacks. Through detailed analyses of triggers and poisoned nodes, we interpret our findings and enhance our understanding of how graph reduction influences robustness against backdoor attacks.
These results highlight the critical need for incorporating robustness considerations in graph reduction for GNN training, ensuring that enhancements in computational efficiency do not compromise the security of GNN systems.
\end{abstract}

\ccsdesc[500]{Security and privacy~Software and application security}
\ccsdesc[500]{Computing methodologies~Machine learning}

\keywords{Trustworthy AI, Graph Backdoor, Graph Neural Network, Graph Reduction, Coarsening, Sparsification}

\received{20 February 2007}
\received[revised]{12 March 2009}
\received[accepted]{5 June 2009}

\maketitle

\section{Introduction}
Graphs are powerful data representations that can be utilized to model entities and their relationships within a wide range of domains such as social networks, biology, recommendation systems and financial networks. 
Graph Neural Networks (GNNs)~\cite{kipf2016semi,hamilton2017inductive,velivckovic2017graph} have recently been widely used on this graph data for various machine learning tasks, including node classification~\cite{xiao2022graph}, graph classification~\cite{errica2019fair} and link prediction~\cite{lei2019gcn}. 
It has been shown to achieve state-of-the-art performance and has found extensive applications such as drug discoveries~\cite{sun2020graph}, traffic forecasting~\cite{wang2020traffic} and personalized recommendation~\cite{weber2019anti}.

Although GNNs have shown success in various domains, they have been shown to be vulnerable to data poisoning attacks~\cite{zügner2018adversarial, zugner2018adversarial,zugner2020adversarial}. By manipulating the dataset in training time through adding/removing edges, perturbing node features, or injecting malicious nodes, these attacks are able to manipulate model prediction and performance. In this paper, we focus on backdoor poisoning attacks~\cite{xi2021graph,zhang2021backdoor,dai2023unnoticeable,zhang2023graph}, which aim to inject a hidden backdoor into the victim GNN model by implanting specific triggers in the training samples. While performing normally on clean inputs, the model trained on poisoned graphs associates the trigger with the target class, leading to misclassification when the trigger is present. As GNNs are progressively utilized in security-critical functions like anomaly detection~\cite{pmlr-v162-tang22b}, malware~\cite{busch2021nf}, and trojans~\cite{yasaei2021gnn4tj}, graph backdoor attacks have posed a significant threat~\cite{alrahis2023graph,alrahis2023poisonedgnn}. Their power and danger lie in their generalizability, affecting any input containing the specific trigger, and their persistent nature, enduring within the model throughout its operational life. This allows attackers to have precise control over the target and the ability of ongoing exploitation.

On the other hand, real-world applications~\cite{ying2018graph,yang2019aligraph} often pose substantial scalability challenges for GNNs, given the vast size of graph data they need to process. For example, the PubMed~\cite{sen2008collective} graph, often cited in research, is considered a large graph with 19,717 nodes, while a snapshot of Twitter social graph in 2010~\cite{kwak2010twitter} consists of millions of users and edges. 
A $K$-layer GNN learns node representations by iteratively aggregating features from a node's $K$-hop neighborhood, which exponentially increases the size of the neighborhood computation graph with each additional layer. This exponential growth causes considerable memory requirements during the training phase, as the mini-batch Stochastic Gradient Descent (SGD) necessitates the loading of large neighborhood computation graphs for each sample in the batch. To address this scalability challenge, graph reduction methods such as coarsening~\cite{huang2021scaling} and sparsification~\cite{hamann2016structure,leskovec2007graph,xu2007scan,satuluri2011local}, which have long been employed in large graph computations, have emerged as promising approaches to reduce the size of computation graphs during GNN training, enabling more scalable GNN training with limited GPU memory~\cite{zhang2023survey,NEURIPS2022_23ee05bf,ijcai2022p772}.

While graph reduction for scalable GNN training has garnered significant interest recently, the impact of these methods on the robustness against backdoor attacks remains largely unexplored. Current graph reduction methods are integrated into GNN training system primarily to improve scalability without considering the presence of state-of-the-art backdoor attacks~\cite{zhang2021backdoor,xi2021graph,dai2023unnoticeable}. However, the coarsening and sparsification processes alter the graph structure, which can inadvertently influence the dynamics of backdoor attacks. These structural changes may disrupt or preserve trigger structures within a reduced graph, potentially mitigating or even exacerbating the attack effectiveness, as demonstrated in our study. The absence of comprehensive studies on this issue highlights a crucial gap in our understanding of the security implications of graph reduction in scalable GNN training. Bridging this gap is vital for developing more robust GNN systems that are not only resilient to adversarial threats but also efficient and scalable—qualities essential for managing large graph data in real-world applications such as social networks, recommendation systems, and anomaly detection.

Therefore, in this paper, we conduct extensive empirical studies to examine the impact of graph reduction on the robustness of GNN against SOTA backdoor attacks~\cite{zhang2021backdoor,xi2021graph,dai2023unnoticeable}. We evaluate the attack success rate (ASR) of three backdoor attacks against GNN training with six graph coarsening and six sparsification methods across various datasets, GNN architectures, graph reduction ratios and attack costs. Additionally, we conduct a detailed analysis of changes in poisoned nodes and triggers to quantify the impact of graph reduction on backdoor attacks, providing in-depth insight into how the reduction can either mitigate or exacerbate attack effectiveness.
To the best of our knowledge, we are the first to investigate the robustness effect of graph reduction for GNN against backdoor attacks. The main contributions of this paper are:
\begin{itemize}[topsep=2pt,leftmargin=*]
    \item \textbf{Mitigation Effect}: We demonstrate the mitigation effect of graph reduction against existing backdoor attacks. Graph reduction can help reduce ASR by more than 10\% to 40\%, which highlights its dual benefits: it improves scalability but also can serve as a viable defense mechanism that significantly weakens the impact of backdoor attacks. We also analyze the impact of multiple factors on the robustness effect of graph reduction, including the attack budget, model architecture, and various reduction methods.
    \item \textbf{Risk of Sparsification}: We find that graph sparsification may inadvertently enhance the effectiveness of backdoor attacks, increasing ASR by approximately 10\% to 20\%. This poses a security concern for scalable GNN training systems that use sparsification as a reduction strategy.
    \item \textbf{Node Level Analysis}: We provide in-depth insights into the robustness effects of graph reduction by quantifying the resulting changes of triggers due to graph coarsening and sparsification. We further analyze the distribution of successful and failed poisoned nodes, demonstrating that graph coarsening effectively mitigates vulnerabilities in low-degree nodes, while sparsification does not.
\end{itemize}
Our results and analysis advocate for a more security-aware approach in the application of graph reduction techniques for GNN training. This approach will ensure that enhancements in computational efficiency and scalability do not compromise the security of GNN systems, particularly when facing sophisticated backdoor attacks. Such considerations are essential for developing GNN applications that are both efficient and secure in real-world scenarios.
\section{Background and Related Work}
\subsection{Graph neural network (GNN)}
Graph Neural Networks (GNN) have emerged as a powerful framework for learning graph representations that capture node features and graph topology. Typical GNN models follow a message-passing framework proposed by Gilmer et al. ~\cite{gilmer2017neural}, where nodes iteratively exchange information with their immediate neighbors across multiple iterations or layers. At each layer, each node aggregates messages from its adjacent nodes and uses the aggregated information to update node representations. GNNs are effectively applied to both \emph{node classification} and \emph{graph classification} tasks. In node classification, GNNs predict the labels of individual nodes using their specific node representations. For graph classification, GNNs typically aggregate the representations of all nodes within a graph to form a signal graph-level representation,  which is then used to determine the label of the entire graph.

Various GNN architectures have been developed to enhance the capabilities of graph representation learning. Kipf et al.~\cite{kipf2016semi} introduced Graph Convolutional Networks (GCN), which extend the concept of convolution to graph data. Attention mechanisms have also been integrated into GNNs, exemplified by the Graph Attention Network (GAT) proposed by Veli{\v{c}}kovi{\'c} et al.~\cite{velivckovic2017graph}. GAT leverages the self-attention of neighbor nodes for the aggregation. By weighting the contributions of neighbors, it enhances model flexibility and expressiveness without being solely dependent on the graph structure. GraphSAGE, proposed by Hamilton et al.~\cite{hamilton2017inductive}, represents another advancement by sampling fixed-size neighborhoods to learn node embeddings, enabling GNNs to scale to large graphs efficiently. 

\subsection{Graph Reduction}
GNN reduction is a data-preprocessing approach that has long been utilized to address the scalability of large graph computation and storage efficiency by reducing graph size. Recently, it has been exploited as a promising solution for accelerating GNN training~\cite{zhang2023survey,ijcai2022p772}.
Typically, there are two types of graph reduction methods: graph coarsening and sparsification.

\noindent\textbf{Graph Coarsening}: Graph coarsening is a technique that reduces the size of a graph while preserving its structural properties. It is applied to large graphs to reduce computational complexity. Given a graph $G$, this approach produces a coarsened graph $G'$ by merging $G$'s nodes in the same local clusters into a ``super-node'' and merges $G$'s edges connecting two super-nodes to a ``super-edge''. $G'$ thus has fewer nodes and edges than $G$. 
To preserve the graph spectrum, previous works~\cite{loukas2018spectrally},\cite{loukas2019graph} proposed coarsening algorithms with spectral approximation guarantees. In addition, there are coarsening approaches that aim to preserve different graph properties, such as electrical properties~\cite{dorfler2012kron}  and connectivity information~\cite{ron2011relaxation}.
Recent work~\cite{huang2021scaling} proposed applying graph coarsening for scalable GNN training, showing a significant reduction in memory costs without a noticeable accuracy loss.

\noindent\textbf{Graph Sparsification}: Graph sparsification is a natural approach to accelerate GNN training by removing less-important edges while preserving all nodes. It typically aims to find a sparsified $G'$ that can well approximate the properties of the original graph $G$. Existing methods typically preserve graph structural properties by selecting edges to retain based on criteria such as node degree~\cite{hamann2016structure} and node similarity~\cite{xu2007scan,satuluri2011local} or by relying on probabilistic methods~\cite{leskovec2007graph}.


\subsection{Graph Poisoning and Backdoor Attack}
Graph poisoning attacks aim to degrade the performance of GNNs or mislead the model by modifying the graph structure (e.g., by adding or removing edges) or node features during training time. There are two types of attacks: 1) targeted attacks~\cite{zügner2018adversarial, zugner2018adversarial}, which aim to misclassify a specific target node; 2) non-targeted attacks~\cite{liu2019unified,zugner2020adversarial}, which aim to degrade the overall model performance.
Backdoor poisoning attack is a special type of targeted poisoning attack, designed to compromise GNN models by associating a particular trigger pattern with a target class.
During the training phase, this attack involves attaching a trigger to a poisoned node and altering its label. This attack ensures that the model performs normally on clean inputs but misclassifies any input containing the trigger as belonging to the target class. 
\textbf{SBA} (Subgraph Backdoor Attack)~\cite{zhang2021backdoor} is a subgraph-based backdoor attack for graph classification. It generates random subgraphs as universal triggers using the Erd\"{o}s-R\'{e}nyi (ER) model and implants the subgraph trigger into multiple locations in the target graph. It has two variants, \textbf{SBA-Samp} and \textbf{SBA-Gen}~\cite{zhang2021backdoor}, which differ in their node feature generation. SBA-Samp generates node features for triggers by randomly sampling from the training graph, whereas SBA-Gen generates them from a Gaussian distribution.
\textbf{GTA} (Graph Trojan Attack)~\cite{xi2021graph} adaptively generates subgraph triggers for different samples by using neural networks. To enhance evasiveness, this attack involves embedding a trigger into the target graph by identifying a subgraph within it that is similar to the trigger, and then replacing the similar subgraph with the trigger. It applies to both graph and node classification tasks.
A recent attack~\textbf{UGBA} (Unnoticeable Graph Backdoor Attacks)~\cite{dai2023unnoticeable} utilizes node representation clustering to determine crucial representative nodes to be poisoned against node classification. It employs an unnoticeable constraint to generate subgraph triggers, which ensures the feature similarity among trigger nodes and poisoned nodes, effectively countering prune-based defenses.



\section{Methodology}

\subsection{System Assumption}
In this paper, we focus on node classification tasks in the inductive setting, which are widely applied in the real world and are particularly relevant for applications involving large-scale graphs~\cite{ying2018graph,yang2019aligraph}. We consider two primary reduction methods: graph coarsening and sparsification. A graph is denoted by $G = (V, E, X)$ where $V$ is the set of vertices, $E$ is the set of edges, and $X$ is the feature matrix (i.e., $X$'s $i$-th row is the feature vector of node $v_i \in V$). Using graph reduction methods, graph acceleration system can be represented by a function $f(G)=G'$ where $G'=(V', E', X')$ is a coarsened or sparsified graph transformed from $G$.

For graph coarsening, we follow the framework proposed in~\cite{huang2021scaling}. The coarsened graph $G'$ has $|V'| < |V|$ and $|E'|<|E|$. It is obtained from $G$ by first computing a partition \(P = \{C_1, C_2, \ldots, C_{n}\}\) of \(V\), i.e., the clusters \(C_1 \ldots C_{n}\) are disjoint and cover all the nodes in \(V\). 
Each cluster \(C_i\) becomes a ``super-node'' in \(G'\) and the ``super-edge'' between two super-nodes \(C_i, C_j\) has weight equal to the total number of edges connecting nodes in between \(C_i\) and \(C_j\). Let $\hat{P}$ be the partition matrix where $\hat{P}_{ij}=1$ if vertex $i$ belongs to cluster $C_j$ otherwise 0. A normalized partition matrix $P=\hat{P}C^{-1/2}$ where $C$ is a diagonal matrix with entries $C_{ii}$ is the number of vertices in $C_i$.
Accordingly, $X'= P^T X$. For node classification, the node labels of $G'$ is computed by $Y' = \arg\max(P^T Y)$, which means that a super node's label is the dominating label in the cluster. There are various coarsening methods to determine the partition $P$, each with different objectives such as preserving structural integrity~\cite{karypis1998fast} or spectral properties~\cite{dorfler2012kron}.

For graph sparsification~\cite{hamann2016structure,xu2007scan,satuluri2011local,leskovec2007graph}, the sparsified graph $G'$ has $V'\subseteq V$, $X'_{V'}=X_{V'}$ and $|E'|<|E|$. It is obtained from $G$ by removing unimportant edges based on specific metrics designed to preserve the structural properties of the graph. The features and labels remain unchanged for the nodes retained by the sparsification. Different sparsification methods employ various strategies to identify unimportant edges, using approaches such as random selection, degree-based selection~\cite{hamann2016structure}, or node similarity~\cite{xu2007scan}.

For a graph coarsening method, we define \textbf{coarsening ratio} $c$ the ratio of the number of nodes in the coarsened graph $G'$ to the number of nodes in the original graph $G$, i.e., $c = |V'|/|V|$. 
Similarly, the \textbf{sparsification ratio} $s$ is defined as the ratio of the number of edges in the sparsified graph $G'$ to the number of edges in $G$, i.e., $s = |E'|/|E|$. We refer to both as reduction ratios in a general context.

\subsection{Attack Model}
\begin{figure}
\centering
\includegraphics[width=0.5\textwidth]{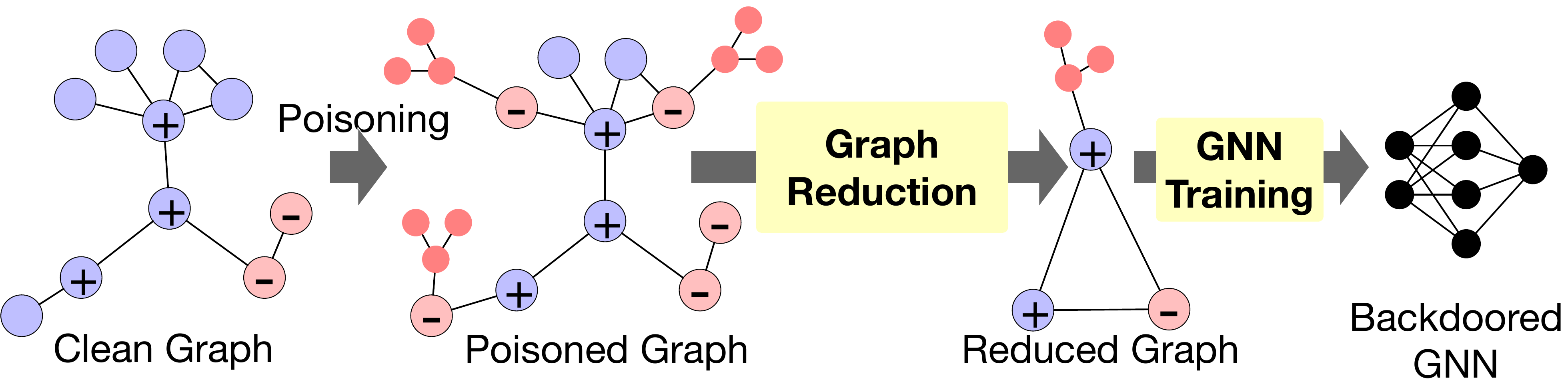}
\caption{Backdoor attack on GNN under graph reduction.}
\label{fig: overall methods}
\vspace{-10pt}
\end{figure}

As the settings of the existing graph backdoor attacks~\cite{zhang2021backdoor,xi2021graph,dai2023unnoticeable}, it is assumed that the adversary can only access the training data and poison certain training samples. The adversary has no control over graph reduction and GNN training systems and has no knowledge about the target GNN models. They are capable of attaching triggers and labels to nodes in the training graph. These same triggers can also be injected into test graphs during inference time to manipulate the model prediction. Figure \ref{fig: overall methods} illustrates the attack scenario under graph reduction.

\noindent\textbf{Graph Backdoor Parameters}: 
A graph backdoor attack uses adaptive subgraphs as triggers, which are injected into the training graph. It involves selecting $n_p$ nodes from the graph as poisoned nodes. A subgraph trigger is then attached to each of these nodes through an edge, and their labels are changed to a target label chosen by the attacker. Thus, a backdoor attack involves three types of parameters:
\begin{itemize}[topsep=1pt]
    \item \textbf{Trigger size}: The number of nodes in a trigger is regarded as trigger size. We denote trigger size as $t$.
    \item \textbf{Poisoning ratio}: During the training stage, $n_p$ nodes get poisoned. Poisoning ratio is the fraction of the number of poisoned nodes to the total number of nodes $n$ in the original graph. We denote the poisoning ratio as $\rho = n_p/n$.
    \item \textbf{Trigger synthesis method}: This involves generating subgraph triggers of a specified size $t$  using various attack methods. In this paper, we consider three state-of-the-art graph backdoor attacks, SBA~\cite{zhang2021backdoor}, GTA~\cite{xi2021graph} and UGBA~\cite{dai2023unnoticeable}.
\end{itemize}

The attack budget or cost is determined by the poisoning ratio and trigger size. The adversary operates under a limited attack budget, because the injection process can be costly, particularly in scenarios like social networks involving fake accounts and social engineering. Additionally, extensive modifications to large graphs can be easily detected.

\subsection{Evalution Framework}
\label{ssec:evaluationframework}
The evaluation of the effectiveness of backdoor attacks typically involves two metrics: \textbf{ attack success rate} (ASR), defined as the percentage of target nodes that are successfully predicted as the target class, and \textbf{average clean accuracy} (ACC), defined as the percentage of correct predictions on clean test nodes. Using these metrics, we perform our empirical measurements and studies across three key tasks: robustness effect analysis, trigger analysis, and poisoned node analysis. These tasks collectively aim to comprehensively assess how graph reduction influences the efficacy of backdoor attacks within GNNs.

\noindent\textbf{Robustness Effect Analysis:}~~To understand the impact of graph reduction on robustness, we first establish baselines for ASR and ACC for attacks against GNN training without graph reduction for each backdoor attack. We then measure and compare the ASR and ACC of these attacks under graph reduction as shown in Figure \ref{fig: overall methods}, applying different reduction methods and varying reduction ratios, i.e., coarsening ratio and sparsification ratio. For a comprehensive understanding, we evaluate and compare the effectiveness of attacks under graph reduction by considering the impact of various GNN architectures, trigger sizes, poisoning ratios, and different reduction methods.

Furthermore, to provide an in-depth explanation of our observations and understanding of the interaction between graph reduction and backdoor attacks, we conduct detailed node-level analyses focusing on trigger changes and poisoned nodes.

\noindent\textbf{Trigger Analysis:}~~
The goal of trigger analysis is to quantify the disruption of triggers due to graph reduction to understand how graph reduction affects backdoor attacks. 

For graph coarsening, the trigger nodes may be merged into a super-node with the label of the dominant group in the cluster and features averaged from the original nodes, which could completely dissolve the trigger structure. Therefore, we consider three metrics: merging ratio, label change ratio, and feature distance, to understand the effect of coarsening on backdoor triggers.
\begin{itemize}[topsep=1pt]
    \item \textbf{Merging ratio} ($m$) measures the percentage of triggers that are consolidated into fewer super-nodes than the trigger size $t$ after coarsening. A higher merging ratio suggests that the trigger’s effectiveness may be significantly compromised, as its unique structure is largely lost within the reduced graph.
    \item \textbf{Label change ratio} ($l$) measures the percentage of nodes initially labeled with the target label due to the poisoning attack, that revert to their original labels after coarsening. A higher label change ratio suggests that coarsening effectively mitigates the attack by restoring the original labels, thus weakening the adversary's influence.
    \item  \textbf{Feature distance} ($d$) is calculated as the L2 norm between the average features of the trigger nodes and the attacking (i.e., attaching) nodes. A decrease in feature distance indicates that coarsening has disrupted the feature characteristics of the trigger, potentially reducing its efficacy in executing the attack.
\end{itemize}
For graph sparsification, while the node features and labels are not changed, only the edges are altered.  The removal of edges could disconnect the trigger nodes from the poisoned nodes, potentially neutralizing the attack like Prune. Therefore, we measure \textbf{prune ratio}, defined as the percentage of triggers disconnected from poisoned nodes. However, it might also eliminate benign edges and nodes, making the attack easier within a sparsified neighborhood. To capture this dual effect, we measure \textbf{post-sparsification poisoning ratio}, defined as the ratio of poisoned nodes that remain connected to trigger nodes after sparsification, relative to the size of the sparsified graph.

\noindent\textbf{Poisoned Node Analysis:}~~   
To discern how graph reduction impacts poisoned target nodes differently, we analyze the distribution of nodes where attacks succeeded and failed, focusing on their degree, 2-hop subgraph density, labels, and the impact of different model architectures. By comparing these distributions between scenarios where backdoor attacks occur without graph reduction and scenarios with graph reduction during GNN training, we aim to understand how poisoned models behave differently at testing time. This analysis helps to illuminate the nuanced effects of graph reduction strategies on the robustness of GNNs against backdoor attacks.

\section{Experiment}
\label{sec:robustanalysis}
In this section, we present our empirical results and analysis. We first introduce the experiment setup (Section \ref{ssec:expsetting}) and demonstrate the scalability benefit of graph reduction for GNN (Section \ref{ssec:trainingaccruacy}). Then, we present robustness effect analysis for graph coarsening (Section \ref{ssec:robustcoarsening}) and graph sparsification (Section \ref{ssec:robustsparsification}), trigger analysis (Section \ref{Sec:triggeranalysis}) and poisoned node analysis (Section \ref{Sec: poisoned nodes}).


\subsection{Experimental Settings}
\label{ssec:expsetting}
\noindent\textbf{Datasets:} 
We use four common datasets of different scales for GNN node classification: Cora~\cite{bojchevski2017deep}, Pubmed~\cite{sen2008collective}, DBLP~\cite{bojchevski2018deep} and OGB-arxiv~\cite{hu2020open}. These datasets collectively span a diverse array of domains, including computer science and biomedical research, and vary significantly in size and complexity. Cora, a small-scale dataset, may not necessarily require graph reduction, but we include it for comprehensiveness to provide a broad perspective across different dataset sizes. Detailed statistics of these datasets are provided in Table \ref{table:dataset} in Appendix. 


\noindent\textbf{GNN Models:} Our methodology is agnostic to the specific architecture of GNN classifiers. We demonstrate the universality of our approach using three well-established GNN classifiers, namely, GCN \cite{kipf2016semi}, GraphSAGE \cite{hamilton2017inductive}, and GAT \cite{velivckovic2017graph}.

\noindent\textbf{GNN Reduction Methods:} We use six graph coarsening methods, which include three methods discussed in~\cite{loukas2019graph}: Variation Neighbourhoods (\textit{VN}), Variation Cliques (\textit{VC}), Variation Edges (\textit{VE}), along with three other methods: Heavy Edge Matching (HE)~\cite{loukas2018spectrally}, Algebraic JC~\cite{ron2011relaxation}, and Kron~\cite{dorfler2012kron}. 
For sparsification, we use six common methods, including Random Edge (RE), Random Node Edge (RNE), Local Degree (Degree)~\cite{hamann2016structure}, Local Similarity (Simi)~\cite{satuluri2011local}, Forest Fire (Fire)~\cite{leskovec2007graph}, and Scan\cite{xu2007scan}. A detailed description of these methods can be found in Appendix~\ref{app:methods}.
By default, we present the results using VN as the default method for graph coarsening method and RNE for graph sparsification, unless otherwise specified.

\noindent\textbf{Backdoor Attacks:} We consider four attack methods: GTA~\cite{xi2021graph},  UGBA~\cite{dai2023unnoticeable}, and two SBA variants SBA-Samp and SBA-Gen~\cite{zhang2021backdoor}. We use their open-source implementations for evaluation. UGBA has demonstrated superior performance compared to other backdoor attacks. As reported in~\cite{dai2023unnoticeable}, it can achieve over 90\% attack success rate against two traditional defense strategies, including Prune and Prune+LD, where Prune simply removes edges connecting two nodes with low cosine similarity on their features, and Prune+LD additionally discards the labels of the nodes linked by dissimilar edges. We use UGBA as the default attack due to its superior performance against prune-based defenses.

\noindent\textbf{Evaluation:} Our evaluation protocol follows the previous work~\cite{dai2023unnoticeable}. We randomly select 10\% of the dataset to be used as target nodes for attack performance evaluation and 10\% as clean test nodes to evaluate clean accuracy, i.e., the prediction accuracy of backdoored models on normal samples. The remaining 80\% of nodes will be used as training graph $G$, including 20\% as the labeled training node set and 10\% as validation set, and 50\% unlabeled nodes. Experiments on each target GNN architecture were conducted five times. We report the average results of backdooring three GNN architectures, resulting in a total of 15 runs.




\subsection{GNN Training with Graph Reduction}
\label{ssec:trainingaccruacy}
We first validate that graph reduction significantly improves scalability while only causing minimal accuracy loss in our experiment scenarios. The results demonstrate the substantial benefits of graph reduction for processing large-scale graphs and establish a clear baseline from which to evaluate robustness.

\begin{table}
\centering
\caption{Accuracy (\%) of Different Coarsening Methods with a Coarsening Ratio of 30\%$|$50\%$|$70\%$|$90\% Respectively.}
\label{tab:coarseningacc}
\scriptsize
\begin{tabular}{l|ccc} 
\toprule
  Dataset (Orig.Acc)             & VN                                                   & VE                                                   & VC                                                    \\ 
\hline
Cora (83.3)      & 75.6$|$78.9$|$81.8$|$83.5 & 71.9$|$80.0$|$82.7$|$83.9 & 69.0$|$80.3$|$83.1$|$83.3  \\
Pubmed (84.9)    & 84.3$|$84.5$|$84.9$|$84.9 & 82.8$|$84.0$|$84.8$|$84.9 & 82.9$|$84.1$|$84.6$|$85.0  \\
DBLP (84.1)      & 79.7$|$80.8$|$82.8$|$84.0 & 75.1$|$81.6$|$83.0$|$83.9 & 76.6$|$81.6$|$83.2$|$83.7  \\
OGB-arxiv (64.1) & 61.9$|$63.3$|$63.9$|$64.1                                  & 61.3$|$63.4$|$64.3$|$64.3                                  & 61.6$|$63.3$|$64.3$|$64.4                                   \\ 
\hline
               & JC                                                   & HE                                                & Kron                                                  \\ 
\hline
Cora (83.3)      & 72.9$|$81.7$|$83.5$|$83.0 & 74.2$|$80.4$|$83.9$|$84.4 & 72.6$|$82.2$|$83.9$|$84.4  \\
Pubmed (84.9)    & 83.9$|$84.4$|$84.7$|$84.9 & 83.1$|$84.1$|$84.9$|$84.9 & 84.0$|$84.5$|$84.8$|$84.9  \\
DBLP (84.1)      & 79.5$|$82.3$|$83.6$|$83.8 & 76.1$|$81.8$|$83.3$|$83.9 & 73.1$|$80.9$|$83.4$|$83.9  \\
OGB-arxiv (64.1) & 62.1$|$63.7$|$64.1$|$64.4                                  & 61.6$|$63.8$|$64.3$|$64.3                                  & 61.9$|$64.2$|$64.3$|$64.3                                   \\
\bottomrule
\end{tabular}
\label{Tab. ACC Coarsening Methods}
\end{table}

\begin{table}
\centering
\caption{Accuracy (\%) of Different Sparsification Methods with a Sparsification Ratio of 30\%$|$50\%$|$70\%$|$90\% Respectively.}
\label{tab:sparsificationacc}
\scriptsize
\begin{tabular}{l|ccc} 
\toprule
Dataset (Orig.Acc)                & RNE                                                  & RE                                                   & Simi                                                  \\ 
\hline
Cora (83.3)      & 80.2$|$81.9$|$82.7$|$82.9 & 80.8$|$83.4$|$83.8$|$83.6 & 82.7$|$83.1$|$84.0$|$83.9  \\
Pubmed (84.9)    & 85.0$|$85.0$|$85.1$|$85.1 & 85.2$|$84.7$|$85.2$|$85.0 & 84.9$|$84.8$|$85.0$|$85.0  \\
DBLP (84.1)      & 83.1$|$83.3$|$83.4$|$83.9 & 83.3$|$83.7$|$84.0$|$84.1 & 83.8$|$84.0$|$83.9$|$83.8  \\
OGB-arxiv (64.1) & 55.4$|$59.2$|$63.1$|$64.1                                  &            60.7$|$62.5$|$63.5$|$64.0                                          &    62.4$|$63.3$|$63.8$|$64.0                                                \\ 
\hline
               & Degree                                               & Forest Fire                                                 & Scan                                                  \\ 
\hline
Cora (83.3)      & 80.8$|$80.9$|$81.8$|$83.5 & 79.6$|$80.8$|$82.3$|$83.0 & 81.0$|$83.1$|$82.9$|$83.4  \\
Pubmed (84.9)    & 85.0$|$85.1$|$85.1$|$85.0 & 85.3$|$85.3$|$85.3$|$85.2 & 84.9$|$85.0$|$84.9$|$84.9  \\
DBLP (84.1)      & 83.7$|$83.9$|$83.9$|$84.1 & 83.2$|$83.5$|$83.6$|$83.7 & 83.0$|$83.9$|$84.0$|$84.1  \\
OGB-arxiv (64.1) &   62.4$|$63.2$|$63.7$|$64.0                                                   &      61.1$|$62.8$|$63.6$|$63.8                                                &                              61.0$|$62.7$|$63.6$|$64.1                          \\
\bottomrule
\end{tabular}
\label{Tab. ACC Sparsification Methods}
\end{table}

We evaluated model accuracy during training across three GNN models (GCN, GAT, GraphSAGE), in a benign setting without adversaries. The results are provided in Table \ref{tab:coarseningacc} and \ref{tab:sparsificationacc}. We observed that, in general, within a wide range of reduction ratios, graph reduction causes only slight accuracy loss compared to the original model accuracy trained without reduction indicated by `Orig.ACC'. For example, on Pubmed, with a ratio 50\% for both coarsening and sparsification, the accuracy changes only slightly, between -0.9\% and +0.4\%. However, there are exceptions; for Cora, due to its small scale, it has an accuracy drop of up to 7\% when $c=0.3$, while OGB-arxiv, the largest one, even up to a 9\% drop in accuracy at a low sparsification ratio $s=0.3$. Nonetheless, by selecting appropriate reduction ratios, we can generally minimize the impact on accuracy within 2\% drop, while significantly reducing memory costs. We measured memory utilization during GNN training, and notably, for large datasets Physics and OGB-arxiv, graph reduction with $c=0.3$ or $s=0.3$ reduces memory consumption by more than 50\%. Further details are provided in Appendix Table \ref{Fig. Memory Coarsening} and \ref{Fig. Memory Sparsification} for coarsening and sparsification, respectively.

\subsection{Robustness Effect Analysis for Coarsening}
\label{ssec:robustcoarsening}
In this section, we present the results of robustness effect of graph coarsening against backdoor attacks.

\subsubsection{Attack mitigation with graph coarsening}
Table \ref{tab:coarsening_comparision} shows our experiment results on four datasets using VN with various coarsening ratios under different attacks (GTA, UGBA, SBA-Samp and SBA-Gen) with a fixed attack budget (trigger size 3 and poisoning ratio 5\%). The reason for this choice of attack budget is that UGBA has been shown to be highly effective with small trigger size $t=3$ and poisoning ratio $\rho=5\%$ while maintaining a minimal clean accuracy drop for being stealthy, as shown in our Table \ref{table:accuracy_comparison} in the later section. Therefore, we use this setting as default. Furthermore, we evaluated two defense mechanisms from previous work~\cite{dai2023unnoticeable} to demonstrate the comparative or stronger attack mitigation effects of graph coarsening compared to these existing defenses.

\begin{table}
\centering
\scriptsize
\setlength{\tabcolsep}{3.5pt}
\caption{ ASR (\%)$|$ACC (\%) results with attack setting $t=3$, $\rho=5\%$. Methods that keep the accuracy drop within 2\% while also having the lowest ASR are highlighted in \textbf{bold}.}
\label{tab:coarsening_comparision}
\begin{tabular}{lcccccc}
\toprule
\textbf{Datasets} & \textbf{Defense} & \begin{tabular}[c]{@{}c@{}} \textbf{Clean} \\ \textbf{Graph}\end{tabular} & \textbf{GTA} & \textbf{UGBA} & \textbf{SBA-Samp} & \textbf{SBA-Gen} \\
\midrule
\multirow{7}{*}{Cora} 
& None      & 83.09 & 67.25$|$83.88          & 93.38$|$85.25            &  39.70$|$86.66           &  58.25$|$86.41 \\
& Prune     & --    & 44.28$|$72.51          & 91.39$|$78.66            &  \textbf{18.94$|$85.56}  &  29.40$|$85.56  \\
& Prune+LD  & --    & 33.55$|$78.25          & 93.30$|$78.90            &  \textbf{18.94$|$85.56}  & \textbf{29.27$|$85.65} \\
& c=0.9     & --    & \textbf{63.20$|$81.51} & 88.95$|$85.08            & 32.26$|$85.87            &  54.98$|$86.27  \\
& c=0.7     & --    & 51.33$|$78.88          & 67.87$|$83.80            & 37.34$|$84.91            & 53.11$|$84.22 \\
& c=0.5     & --    & 43.64$|$72.29          & \textbf{49.84$|$81.01}   &  31.88$|$84.96           & 43.03$|$83.13 \\
& c=0.3     & --    & 43.61$|$65.67          & 35.76$|$75.29            &  29.17$|$81.80           & 39.80$|$82.96 \\
\midrule
\multirow{7}{*}{Pubmed} 
& None      & 86.86 & 85.96$|$84.93          & 83.07$|$85.67            & 32.14$|$84.74            & 24.97$|$86.00 \\
& Prune     & --    & 92.21$|$85.55          & 81.81$|$84.88            & 22.36$|$86.08            & 20.77$|$86.08 \\
& Prune+LD  & --    & 73.25$|$85.99          & 82.62$|$85.21            & \textbf{22.31$|$86.05}   & \textbf{20.69$|$86.07} \\
& c=0.9     & --    & 67.46$|$84.91          & 73.97$|$85.51            & 33.02$|$84.95            &  25.22$|$86.07  \\
& c=0.7     & --    & 48.23$|$83.95          & 66.36$|$85.31            & 28.22$|$83.52            &  25.55$|$85.83 \\
& c=0.5     & --    & 45.67$|$83.82          & \textbf{57.32$|$85.21}   & 26.32$|$83.20            &  26.42$|$85.73 \\
& c=0.3     & --    & \textbf{44.43$|$83.13} & 56.77$|$84.46            & 24.61$|$80.86            & 27.93$|$84.85 \\
\midrule
\multirow{7}{*}{DBLP} 
& None      & 84.40 & 76.00$|$84.07          & 80.26$|$84.38            & 32.14$|$84.74            &  42.24$|$84.75  \\
& Prune     & --    & 96.88$|$82.72          & 82.36$|$82.57            & 15.65$|$83.95            &  15.08$|$84.11  \\
& Prune+LD  & --    & 90.40$|$82.41          & 87.54$|$82.50            & \textbf{15.43$|$84.11}   & \textbf{15.05$|$84.11}  \\
& c=0.9     & --    & \textbf{64.21$|$83.34} & 75.17$|$84.43            & 33.02$|$84.95            &  41.32$|$84.76  \\
& c=0.7     & --    & 59.64$|$81.27          & \textbf{70.76$|$83.45}   & 28.22$|$83.52            &  37.19$|$83.97  \\
& c=0.5     & --    & 56.32$|$78.54          & 63.54$|$80.88            & 26.32$|$82.00            &  38.36$|$82.22  \\
& c=0.3     & --    & 53.62$|$74.32          & 62.84$|$77.98            & 24.61$|$80.86            &  34.04$|$80.51  \\
\midrule
\multirow{7}{*}{\begin{tabular}[c]{@{}c@{}}OGB\\-arxiv\end{tabular}} 
& None      & 65.50 & 37.02$|$59.86          & 73.35$|$63.84            & 0.03$|$65.71              & 0.02$|$65.84 \\
& Prune     & --    &  0.16$|$62.46          & 72.07$|$61.58            & \textbf{0.01$|$66.06}& 0.03$|$65.89  \\
& Prune+LD  & --    &  0.28$|$62.46          & 66.95$|$62.92            & 0.01$|$66.12              & 0.02$|$65.89  \\
& c=0.9     & --    & 34.06$|$52.13          & \textbf{63.54$|$63.09}            & 0.03$|$65.70              & 0.02$|$65.83 \\
& c=0.7     & --    & 37.01$|$55.53          & 49.96$|$61.69            & 0.05$|$64.80              & 0.02$|$65.27 \\
& c=0.5     & --    & 29.45$|$46.86          & 45.83$|$61.25            & 0.05$|$64.62              & 0.03$|$64.50 \\
& c=0.3     & --    & 14.13$|$39.72          & 28.43$|$60.54            & 0.04$|$64.04              & \textbf{0.00$|$64.08}\\
\bottomrule
\end{tabular}
\vspace{-10pt}
\end{table}

We focus on graph coarsening ratios that ensure an accuracy drop within 2\%. From the table, we can see that in these cases, ASRs of GTA and UGBA are effectively reduced by up to 40\% for large datasets. For example, on Pubmed, with a coarsening ratio $c=0.3$, the ASR for GTA drops from 85.96\% to 44.43\%, and for UGBA, the ASR decreases from 83.07\% to 57.32\%. For the small dataset Cora, graph coarsening reduces the ASR of GTA from 67.25\% to 63.20\% at $c=0.9$, and the ASR of UGBA from 93.38\% to 49.84\% at $c=0.5$. Moreover, graph coarsening is shown to mitigate the ASRs of GTA and UGBA more effectively than two traditional defenses, including Prune and Prune+LD. In particular, the previous work~\cite{dai2023unnoticeable} reported that UGBA achieves over 90\% attack success
rate against these two defenses. Our study shows that with graph coarsening, UGBA's ASR can be reduced by 10\% up to 40\%.

In contrast, the Prune and Prune+LD methods achieve a better mitigation effect on the SBA-Samp/Gen attack than graph coarsening. This is because, in the SBA method, trigger node features are either randomly generated based on the statistical properties (mean and standard deviation) of the original features or directly sampled from the original feature set. The Prune and Prune+LD methods, which remove triggers based on node feature similarity, are more effective under these conditions. The random feature generation in SBA makes it easier for the Prune and Prune+LD methods to detect the feature discrepancies between the triggers and the attacking nodes, thus allowing for more targeted removal of injected triggers.

In summary, the effectiveness of graph coarsening in reducing ASR varies across different attack methodologies. Notably, coarsening significantly diminishes the ASR of the state-of-the-art attack UGBA, which typically exhibits high success rates against Prune-based defenses. This impact of coarsening extends across various datasets, proving particularly potent in large-scale datasets such as Pubmed, DBLP and OGB-arxiv, where, in some instances, its defensive capabilities surpass traditional methods like Prune. Overall, our experimental results demonstrate \textbf{the dual benefit of coarsening that improves the scalability of GNN training for large-scale graphs and concurrently enhances their resilience to backdoor attacks}.

\subsubsection{ASR and ACC v.s. Coarsening ratio}
\begin{figure}
    \centering
        \begin{subfigure}[b]{0.2\textwidth}
        \centering
        \includegraphics[width=\textwidth,height=2.8cm]{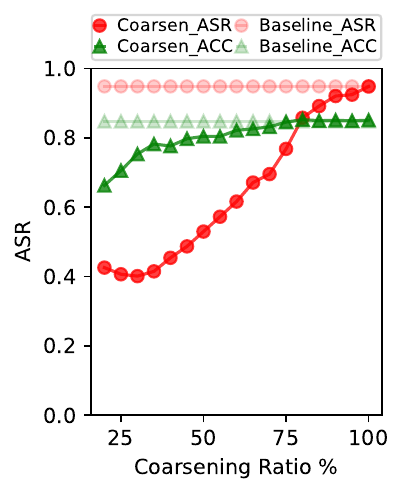}
        \caption{Cora}
        \label{Fig: acc_asr_Cora}
    \end{subfigure}
    \begin{subfigure}[b]{0.2\textwidth}
        \centering
        \includegraphics[width=\textwidth,height=2.8cm]{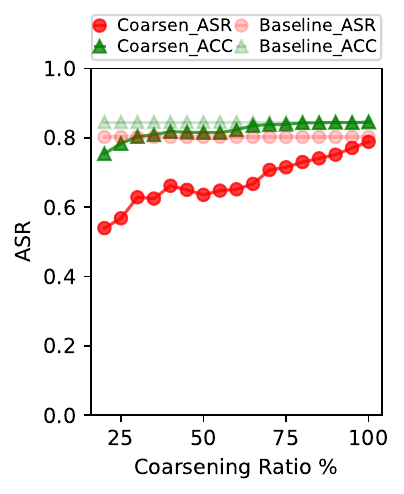} 
        \caption{DBLP}
        \label{Fig: acc_asr_DBLP}
    \end{subfigure}
    \begin{subfigure}[b]{0.2\textwidth}
        \centering
        \includegraphics[width=\textwidth,height=2.8cm]{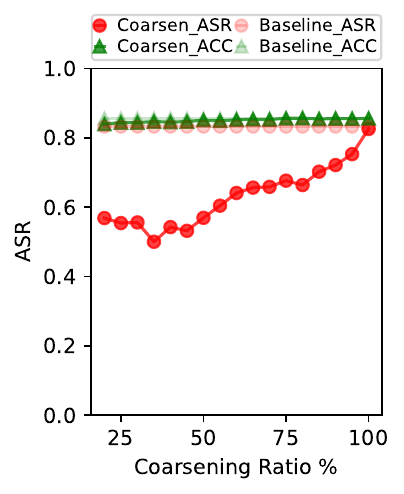}
        \caption{Pubmed}
        \label{Fig: acc_asr_Pubmed}
    \end{subfigure}
    \begin{subfigure}[b]{0.2\textwidth}
        \centering
        \includegraphics[width=\textwidth,height=2.8cm]{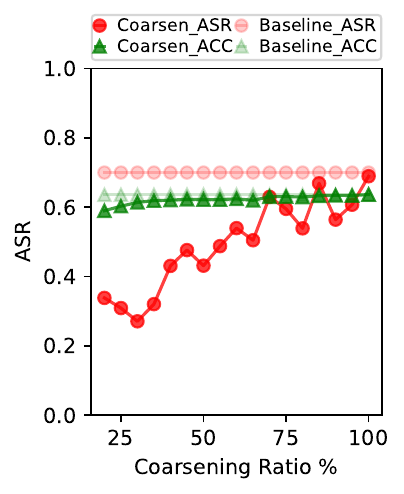} 
        \caption{OGB-arxiv}
        \label{Fig: acc_asr_ogbn}
    \end{subfigure}
    \caption{Impact of Graph Coarsening on ASR with Trigger Size 3 and Poisoning Ratio 5\%.}
\label{Fig. acc asr}
\vspace{-10pt}
\end{figure}

To better understand how the coarsening ratio affects ASR and ACC for existing backdoor attacks, we show the ASR and ACC of UGBA under various coarsening ratios for four datasets in Figure \ref{Fig. acc asr}. `Baseline\_ACC' and `Baseline\_ASR' represent UGBA's ACC and ASR without coarsening poisoned training dataset, while `Coarsen\_ACC' and `Coarsen\_ASR' refer to the ACC and ASR when coarsening is applied. 
With coarsening, we can see a significant decrease in ASR compared to Baseline\_ASR, especially as the coarsening ratio decreases. Meanwhile, Coarsen\_ACC remains relatively constant regardless of changes in the coarsening ratio $c$. For example, within only a 2\% accuracy drop from Baseline\_ACC, Pubmed's ASR decreases by 30\% at $c=45\%$ and OGB-arxiv's ASR decreases by 37\% at $c=25\%$. These results indicate the effective mitigation of graph coarsening against backdoor attacks, suggesting its viability as a layer of defense.

\subsubsection{Impact of trigger size and poisoning ratio}
\label{Sec: trigger and size}
To understand how the robustness effect of graph coarsening changes with backdoor attack cost, we further tested UGBA with different poisoning ratios (5\%, 10\%, 15\%) and trigger sizes (3, 6, 9) under different coarsening ratios.

Table \ref{table:accuracy_comparison} shows the ACC results. The `$c$(ACC\%)' column indicates the baseline accuracy of the model trained without poisoning, under different coarsening ratios.
It is clear that a higher poisoning ratio $\rho$ or a larger trigger size $t$ leads to a lower ACC under a given coarsening setting.
Given coarsening ratios that ensure the baseline accuracy loss within 2\% from the no-poisoning no-coarsening case, the clean accuracy under UGBA attack remains close to the baseline accuracy. Within these scenarios, the ACC of UGBA, when using a larger $t$ and a higher $\rho$, tends to be more sensitive to changes in the coarsening ratio on large datasets. For example, for OGB-arxiv with $\rho=10\%$, when c changes from 0.9 to 0.3, ACC for $t=3$ decreases by 5\%, but for $t=9$ it decreases by 27\%, a significantly larger drop. This effect is likely due to the coarsening process consolidating features of injected nodes with original nodes, where a greater number of injected nodes results in more divergence in the feature distribution, leading to substantial accuracy loss.

\begin{table}
\centering
\scriptsize
\caption{ACC (\%) under UGBA attack with a trigger size of 3$|$6$|$9 respectively. `Null': no coarsening performed.}
\begin{tabular}{cr|lll} 
\toprule
\multicolumn{1}{l}{datasets} & \multicolumn{1}{l|}{$c$(ACC\%)}   & \multicolumn{1}{c}{$\rho=5\%$} & \multicolumn{1}{c}{10\%} & \multicolumn{1}{c}{15\%}  \\ 
\hline
\multirow{5}{*}{Cora}    & \multicolumn{1}{l|}{Null(83.1)}& 85.2$|$84.8$|$84.0                 & 84.9$|$81.1$|$85.3       &   83.3$|$80.5$|$60.7      \\
                         & 0.9(86.9)                      & 85.1$|$84.7$|$83.5                 & 84.0$|$80.1$|$84.9       &   83.1$|$78.6$|$63.4     \\
                         & 0.7(85.1)                       & 83.8$|$84.9$|$82.3                 & 78.9$|$77.7$|$81.6       &   77.0$|$72.8$|$63.1     \\
                         & 0.5(83.4)                       & 81.0$|$81.0$|$78.2                 & 73.9$|$72.4$|$77.3       &   71.3$|$62.3$|$48.6     \\
                         & 0.3(79.5)                       & 75.2$|$71.5$|$71.2                 & 64.8$|$70.8$|$72.6       &   59.8$|$46.2$|$38.3     \\
\hline
\multirow{5}{*}{Pubmed}  & \multicolumn{1}{l|}{Null(86.9)}& 85.6$|$85.5$|$85.6                 & 85.5$|$85.4$|$85.5       &   85.2$|$85.1$|$85.3                 \\
                         & 0.9(84.8)                       & 85.5$|$85.4$|$85.4                 & 85.4$|$85.5$|$85.4       &   85.2$|$85.3$|$85.1                 \\
                         & 0.7(84.6)                       & 85.3$|$85.2$|$85.0                 & 85.4$|$85.1$|$85.2       &   84.9$|$85.1$|$85.1                 \\
                         & 0.5(84.5)                       & 85.2$|$84.6$|$84.8                 & 84.7$|$84.4$|$84.6       &   84.2$|$84.4$|$84.5                 \\
                         & 0.3(83.4)                       & 84.4$|$83.5$|$84.1                 & 83.7$|$83.3$|$83.6       &   82.7$|$82.6$|$83.0                 \\ 
\hline
\multirow{5}{*}{DBLP}    & \multicolumn{1}{l|}{Null(84.4)}& 84.3$|$84.5$|$84.6                 & 84.5$|$84.5$|$84.4       &   84.2$|$84.2$|$83.8                 \\
                         & 0.9(84.6)                       & 84.4$|$84.2$|$84.3                 & 84.2$|$84.2$|$84.1       &   83.8$|$83.6$|$83.5                 \\
                         & 0.7(83.7)                       & 83.4$|$83.4$|$84.2                 & 83.5$|$83.3$|$83.2       &   83.1$|$82.8$|$82.2                 \\
                         & 0.5(80.7)                       & 80.8$|$81.9$|$81.3                 & 80.6$|$80.3$|$79.9       &   79.3$|$78.5$|$76.1                 \\
                         & 0.3(77.6)                       & 77.9$|$77.5$|$78.8                 & 78.5$|$77.5$|$77.4       &   76.8$|$76.4$|$71.5                 \\

\hline
\multirow{5}{*}{\begin{tabular}[c]{@{}c@{}}OGB\\-arxiv\end{tabular}}     & \multicolumn{1}{l|}{Null(65.5)} & 63.8$|$61.4$|$61.0                 & 62.9$|$63.2$|$56.4                 &   63.4$|$60.3$|$55.5                 \\
                         & 0.9(64.5)                      & 63.1$|$58.5$|$60.1                 & 62.1$|$63.3$|$49.8                 &   62.8$|$58.8$|$48.8                 \\
                         & 0.7(64.0)                      & 61.7$|$56.2$|$57.1                 & 60.1$|$63.1$|$37.2                 &   61.1$|$53.9$|$38.8                 \\
                         & 0.5(63.6)                      & 61.2$|$53.3$|$54.8                 & 59.5$|$61.9$|$32.1                 &   60.5$|$51.2$|$28.6                 \\
                         & 0.3(63.6)                      & 60.5$|$50.0$|$47.7                 & 57.1$|$57.6$|$22.2                 &   59.3$|$46.5$|$18.6                 \\
\bottomrule
\end{tabular}
\label{table:accuracy_comparison}
\end{table}

Table \ref{Tab: Adversarial Poisoning Ratio} presents ASR results of UGBA with coarsening ratios that ensure clean accuracy drop within 2\% compared to the accuracy in scenarios without poisoning or coarsening. These coarsening ratios include {70\%, 75\%, 80\%, 85\%, 90\%} across three large datasets: PubMed, DBLP, and OGB-arxiv. Under the same coarsening setting, increasing the trigger size from 3 to 9 does not significantly change the ASR. An interesting observation for the PubMed and DBLP datasets is that under coarsening ratios 70\%, 75\%, 80\%, ASR decreases slightly as poisoning ratio $\rho$ increases. A possible explanation is that a higher poisoning ratio introduces a greater number of nodes, which in turn increases the likelihood that trigger nodes will be merged with other nodes when graph coarsening is applied at the same ratio. Table \ref{Tab. UGBA Trigger Analysis Different Rho} in Section \ref{Sec:triggeranalysis} further supports this explanation by showing a slightly higher trigger merging ratio for a higher poisoning ratio on Pubmed.
Nevertheless, it is clear that \textbf{with a higher poisoning ratio, the mitigation effect of coarsening decreases, and a lower coarsening ratio is required to be more effective for reducing ASR}.

\begin{table}
\centering
\scriptsize
\caption{ASR under UGBA attack. `*' denotes the entry that has ACC drop more than 2\% compared to the accuracy on clean data without poisoning or coarsening.}
\label{Tab: Adversarial Poisoning Ratio}
\begin{tabular}{cr|rrr|rrr|rrr} 
\toprule
\multicolumn{1}{l}{}  & \multicolumn{1}{l|}{}  & \multicolumn{3}{c}{Pubmed} & \multicolumn{3}{c}{DBLP} & \multicolumn{3}{c}{OGB-arxiv}                                                 \\
\multicolumn{1}{l}{$t$} & \multicolumn{1}{l|}{$c$} & $\rho$=5\%  & 10\% & 15\%         & 5\%  & 10\% & 15\%       & \multicolumn{1}{r}{5\%} & \multicolumn{1}{r}{10\%} & \multicolumn{1}{r}{15\%}  \\ 
\hline
\multirow{6}{*}{3}    & 70\%                   & 0.67 & 0.6  & 0.56         & 0.63 & 0.69 & 0.66       & 0.48 & 0.50 & 0.64 \\
                      & 75\%                   & 0.68 & 0.62 & 0.60         & 0.65 & 0.69 & 0.67       & 0.46 & 0.47 & 0.55 \\
                      & 80\%                   & 0.68 & 0.64 & 0.67         & 0.66 & 0.73 & 0.70       & 0.49 & 0.50 & 0.57 \\
                      & 85\%                   & 0.70 & 0.67 & 0.70         & 0.68 & 0.76 & 0.73       & 0.56 & 0.50 & 0.60 \\
                      & 90\%                   & 0.72 & 0.73 & 0.74         & 0.70 & 0.76 & 0.74       & 0.51 & 0.54 & 0.61 \\
                      & Null                   & 0.84 & 0.86 & 0.87         & 0.75 & 0.85 & 0.87       & 0.81 & 0.79 & 0.69 \\ 
\hline
\multirow{6}{*}{6}    & 70\%                   & 0.64 & 0.55 & 0.51         & 0.68 & 0.67 & 0.63       & 0.52~* & 0.54~* & 0.59~* \\
                      & 75\%                   & 0.66 & 0.58 & 0.57         & 0.71 & 0.67 & 0.64       & 0.51~* & 0.49~* & 0.50~* \\
                      & 80\%                   & 0.67 & 0.62 & 0.63         & 0.72 & 0.69 & 0.67       & 0.53~* & 0.53~* & 0.57~* \\
                      & 85\%                   & 0.67 & 0.65 & 0.65         & 0.73 & 0.74 & 0.70       & 0.62~* & 0.55~* & 0.58~* \\
                      & 90\%                   & 0.68 & 0.72 & 0.71         & 0.76 & 0.73 & 0.72       & 0.57~* & 0.56~* & 0.58~* \\
                      & Null                   & 0.83 & 0.86 & 0.85         & 0.83 & 0.85 & 0.90       & 0.77~* & 0.77~* & 0.75~* \\
\hline
\multirow{6}{*}{9}    & 70\%                   & 0.55 & 0.54 & 0.52         & 0.63 & 0.68 & 0.63       & 0.51~* & 0.50~* & 0.54~* \\
                      & 75\%                   & 0.60 & 0.57 & 0.56         & 0.66 & 0.69 & 0.64       & 0.51~* & 0.47~* & 0.49~* \\
                      & 80\%                   & 0.62 & 0.60 & 0.62         & 0.68 & 0.71 & 0.67       & 0.52~* & 0.49~* & 0.57~* \\
                      & 85\%                   & 0.63 & 0.63 & 0.65         & 0.68 & 0.75 & 0.70       & 0.61~* & 0.50~* & 0.58~* \\
                      & 90\%                   & 0.65 & 0.70 & 0.71         & 0.71 & 0.75 & 0.72       & 0.56~* & 0.53~* & 0.61~* \\
                      & Null                   & 0.83 & 0.86 & 0.87         & 0.78 & 0.89 & 0.90       & 0.73~* & 0.70~* & 0.80~* \\
\bottomrule
\end{tabular}
\vspace{-10pt}
\end{table}

\subsubsection{Impact of GNN Architectures}
\begin{figure}
    \centering
        \begin{subfigure}[b]{0.18\textwidth}
        \centering
        \includegraphics[width=\textwidth, height=2.5cm]{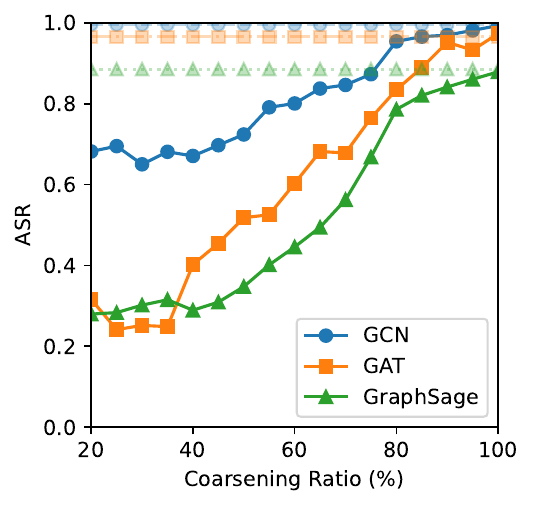} 
        \caption{Cora}
        \label{Fig: asr_cora}
    \end{subfigure}
    \begin{subfigure}[b]{0.18\textwidth}
        \centering
        \includegraphics[width=\textwidth,height=2.5cm]{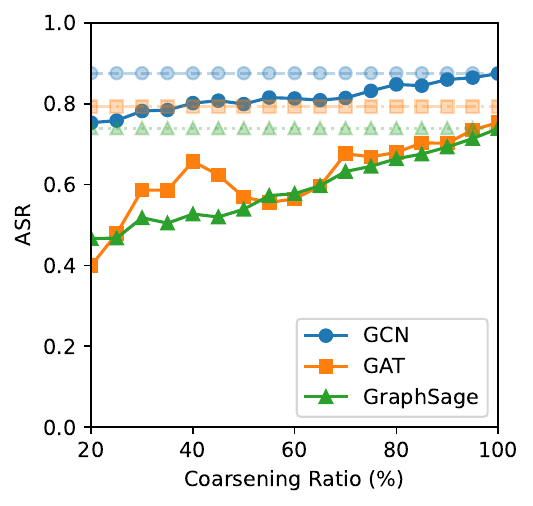} 
        \caption{DBLP}
        \label{Fig: asr_DBLP}
    \end{subfigure}
    \begin{subfigure}[b]{0.18\textwidth}
        \centering
        \includegraphics[width=\textwidth,height=2.5cm]{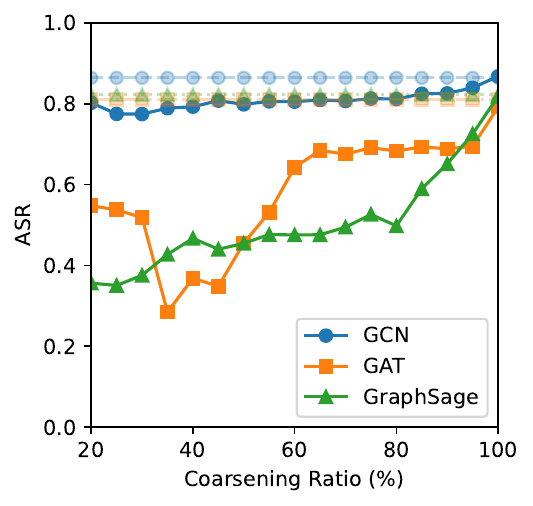} 
        \caption{Pubmed}
        \label{Fig: asr_pubmed}
    \end{subfigure}
    \begin{subfigure}[b]{0.18\textwidth}
        \centering
        \includegraphics[width=\textwidth,height=2.5cm]{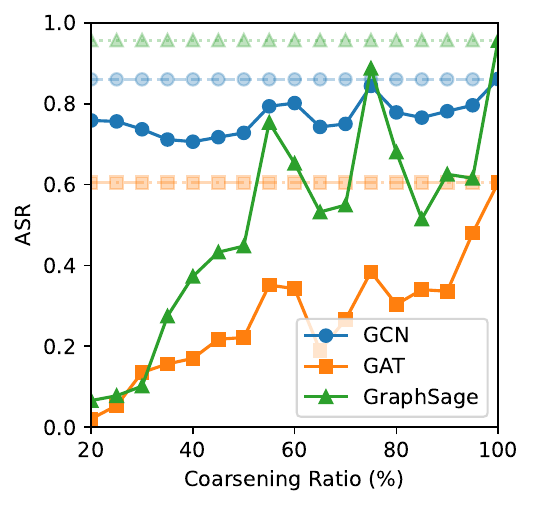} 
        \caption{OGB-arxiv}
        \label{Fig: asr_ogbn}
    \end{subfigure}
    \caption{Impact of Graph Coarsening on ASR with different GNN models. A light-colored line represents the baseline ASR under no coarsening.}
\label{Fig. model asr pubmed ogb}
\vspace{-10pt}
\end{figure}

Although the ACCs remain relatively consistent under coarsening across three different GNN architectures in our experiments, the ASRs vary significantly. Figure \ref{Fig. model asr pubmed ogb} shows the ASR curves under various coarsening ratios on four datasets for GCN, CAT, and GraphSage, respectively. As we can see, the ASRs obtained with GraphSAGE and GAT are lower than those with GCN. For instance, on Pubmed Dataset, when $c=70\%$, the ASR of GCN drops from 86\% to 81\%, while the ASR of GraphSAGE drops from 81\% to 49\% and the ASR of GAT drops from 82\% to 68\%. In the OGB-arxiv dataset, the ASR of GCN drops from 86\% to 75\%, while the ASR of GAT drops from 60\% to 27\% and the ASR of GraphSAGE drops from 96\% to 55\%.

Our result demonstrates that \textbf{graph coarsening mitigates the ASRs for GraphSAGE and GAT more effectively than for GCN}. This discrepancy may be because that GCN aggregates information from all neighboring nodes, while GAT and GraphSAGE sample features from only a subset of neighboring nodes. After coarsening, only some triggers are retained in the graph, making GraphSAGE less likely to aggregate information from trigger nodes and causing GAT to pay less attention to these nodes.

\subsubsection{Impact of different coarsening methods}
\label{Sec: coarsening methods}
\begin{figure}
    \centering
    \centering
    \begin{subfigure}[b]{0.5\textwidth}
    \centering
    \includegraphics[width=\textwidth]{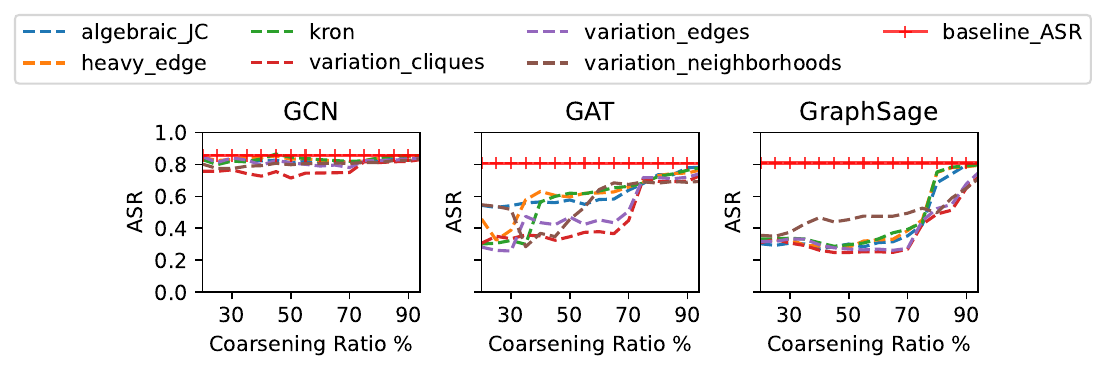}
    \caption{ASR on Pubmed}
    \label{fig:Coarsening Pubmed}
    \end{subfigure}
    \begin{subfigure}[b]{0.5\textwidth}
    \centering
    \includegraphics[width=\textwidth]{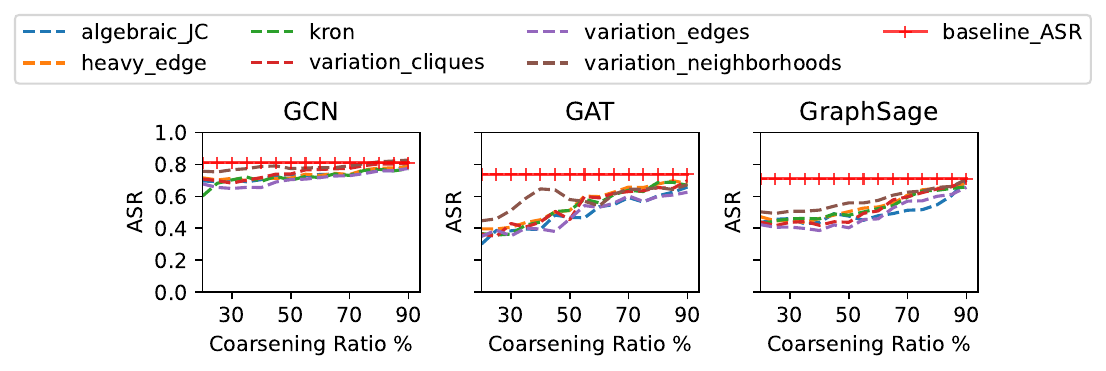}
    \caption{ASR on DBLP}
    \label{fig:Coarsening DBLP}
    \end{subfigure}
    \begin{subfigure}[b]{0.5\textwidth}
    \centering
    \includegraphics[width=\textwidth]{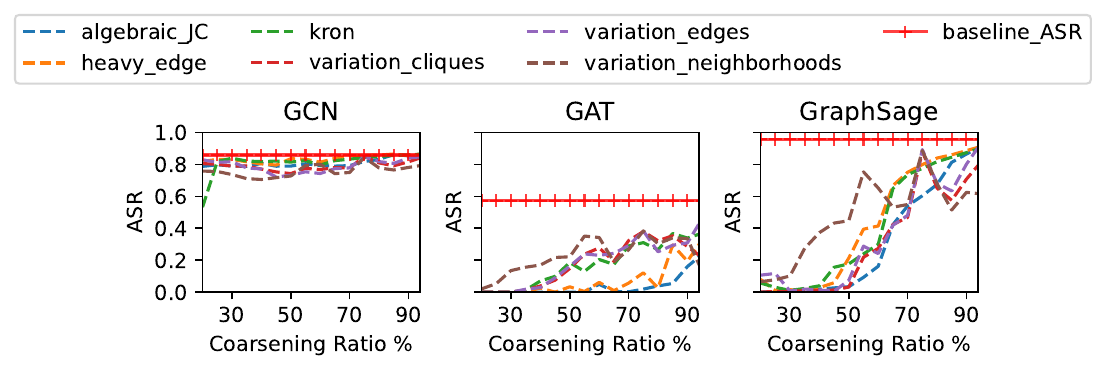}
    \caption{ASR on OGB-arxiv}
    \label{fig:Coarsening OGB}
    \end{subfigure}
        \vspace{-10pt}
    \caption{ASR Under Different Graph Coarsening Methods.}
    \label{fig:asr_vs_coarsening}
        \vspace{-5pt}
\end{figure}

To understand the differences in robustness effects among specific coarsening methods, we tested six methods including \textit{VN}, \textit{VC}, \textit{VE}, Heavy Edge Matching, Algebraic JC, and Kron. Figure \ref{fig:asr_vs_coarsening} presents UGBA's ASR under these different methods on Pubmed, DBLP and OGB-arxiv datasets, under three GNN architectures, respectively. The baseline ASR in the figure is the ASR under no coarsening. 

From the results, we observe that on both datasets, the ASR for the GCN model is only marginally reduced compared to the reductions seen with GAT and GraphSage. \textbf{The impact of different coarsening methods on ASR reduction varies}. The \textit{VC} and \textit{VE} methods have close and consistently overall good performance in mitigating attacks. Specifically, for Pubmed (Figure \ref{fig:Coarsening Pubmed}), \textit{VC} and \textit{VE} outperform others on the GAT model. Unlike the straightforward amalgamation of adjacent nodes, \textit{VC} and \textit{VE} target specific structural components, such as cliques and edges, potentially disrupting trigger structures more effectively. However, on a larger dataset OGB-arxiv, as shown in Figure \ref{fig:Coarsening OGB}, the HE and JC methods outperform others on the GAT model, which may suggest that non-spectrum-based coarsening methods can be more effective at reducing ASR in such large poisoned graphs.

\subsection{Robustness Effect with Graph Sparsification}
\label{ssec:robustsparsification}
This section follows the same methodology to investigate the effect of graph sparsification as in the previous section on graph coarsening. We examine six sparsification methods introduced in Section \ref{ssec:expsetting}. Given that both coarsening and sparsification exhibit similar trends in their effects under different GNN architectures and attack costs, we focus here solely on presenting the ASRs and discussing the unique characteristics of sparsification.

\subsubsection{Results}

\begin{table}
\centering
\scriptsize
\setlength{\tabcolsep}{3.5pt}
\caption{ ASR(\%) $|$ ACC(\%) results with attack setting $t=3$, $\rho=5\%$. Methods that keep the accuracy drop within 2\% while also having the lowest ASR are highlighted in \textbf{bold}.}
\label{tab:sparsification_comparision}
\begin{tabular}{lcccccc}
\toprule
\textbf{Datasets} & \textbf{Defense} & \begin{tabular}[c]{@{}c@{}} \textbf{Clean} \\ \textbf{Graph}\end{tabular} & \textbf{GTA} & \textbf{UGBA} & \textbf{SBA-Samp} & \textbf{SBA-Gen} \\
\midrule
\multirow{7}{*}{Cora} 
& None      & 83.09 & 84.55$|$83.88& 93.38$|$85.25 &  45.76$|$86.66           &  58.25$|$86.41 \\
& Prune     & --    & 44.28$|$72.51 & 91.39$|$78.66 &  \textbf{18.94$|$85.56}  &  29.40$|$85.56  \\
& Prune+LD  & --    & 33.55$|$78.25 & 93.30$|$78.90 &  \textbf{18.94$|$85.56}  & 29.27$|$85.65 \\
& s=0.9     & --    & 99.90$|$84.93 & 93.65$|$84.41 & 47.72$|$87.28& 58.20$|$86.40\\
& s=0.7     & --    &   94.82$|$82.79 & 91.07$|$84.56 & 36.04$|$85.43& 42.95$|$85.31\\
& s=0.5     & --    & 89.83$|$82.93 & 86.93$|$83.01 & 30.14$|$85.30& 37.90$|$85.09\\
& s=0.3     & --    & 86.27$|$79.87 & \textbf{75.10$|$82.17}& 21.53$|$84.32& \textbf{26.69$|$84.59}\\
\midrule
\multirow{7}{*}{Pubmed} 
& None      & 86.86 & 85.96$|$84.93 & 83.07$|$85.67 & 32.14$|$84.74            & 24.97$|$86.00 \\
& Prune     & --    & 92.21$|$85.55 & 81.81$|$84.88 & 22.36$|$86.08            & 20.77$|$86.08 \\
& Prune+LD  & --    & \textbf{73.25$|$85.99}& 82.62$|$85.21 & 22.31$|$86.05   & 20.69$|$86.07 \\
& s=0.9     & --    & 90.09$|$85.49 & 81.61$|$85.48 & 19.00$|$86.10& 19.95$|$86.02\\

& s=0.7     & --    & 91.63$|$85.49 & 76.61$|$85.31 & 16.89$|$86.00& 17.07$|$85.95\\
& s=0.5     & --    & 92.32$|$85.31 & 66.91$|$85.25 & 15.79$|$86.08& 18.21$|$85.88\\
& s=0.3     & --    & 97.65$|$84.25 & \textbf{57.89$|$85.02}& \textbf{14.80$|$85.96}& \textbf{16.13$|$85.95}\\
\midrule
\multirow{7}{*}{DBLP} 
& None      & 84.40 & 91.30$|$84.07& 87.97$|$84.38 & 32.14$|$84.74            &  42.24$|$84.75  \\
& Prune     & --    & 96.88$|$82.72 & 82.36$|$82.57 & 15.65$|$83.95            &  15.08$|$84.11  \\
& Prune+LD  & --    & 90.40$|$82.41 & 87.54$|$82.50 & 15.43$|$84.11   & 15.05$|$84.11  \\
& s=0.9     & --    & 84.39$|$84.17 & \textbf{82.83$|$84.50}& 30.15$|$84.92& 18.51$|$84.52\\
& s=0.7     & --    & 84.98$|$84.02 & 84.40$|$84.34 & 19.31$|$84.38& \textbf{12.51$|$84.43}\\
& s=0.5     & --    & 85.25$|$83.51 & 86.21$|$83.87 & 16.08$|$84.04& 15.27$|$83.96\\
& s=0.3     & --    & 90.31$|$83.30 & 85.94$|$83.39 & \textbf{14.67$|$83.92}& 15.22$|$83.83\\
\midrule
\multirow{7}{*}{\begin{tabular}[c]{@{}c@{}}OGB\\-arxiv\end{tabular}} 
& None      & 65.50 & 60.16$|$59.86 & 73.35$|$63.84 & 0.03$|$65.71            & 0.02$|$65.84 \\
& Prune     & --    &  0.16$|$62.46 & 72.07$|$61.58 & 0.01$|$66.06            & 0.03$|$65.89  \\
& Prune+LD  & --    &  0.28$|$62.46 & 66.95$|$62.92 & \textbf{0.01$|$66.12}& \textbf{0.02$|$65.89}\\
& s=0.9     & --    & 61.06$|$61.46 & 86.04$|$62.42 & 0.02$|$65.03& 0.05$|$65.06\\
& s=0.7     & --    & 63.16$|$60.44 & 90.14$|$60.73 & 0.00$|$62.52& 0.00$|$62.53\\
& s=0.5     & --    & 64.15$|$58.24 & 75.14$|$58.02 & 0.00$|$59.64& 0.00$|$59.65\\
& s=0.3     & --    & 66.86$|$51.51 & 48.42$|$54.51 & 0.00$|$55.60& 0.00$|$55.92\\
\bottomrule
\end{tabular}
\end{table}

\begin{figure}
    \centering
    \begin{subfigure}[b]{0.5\textwidth}
    \centering
    \includegraphics[width=\textwidth]{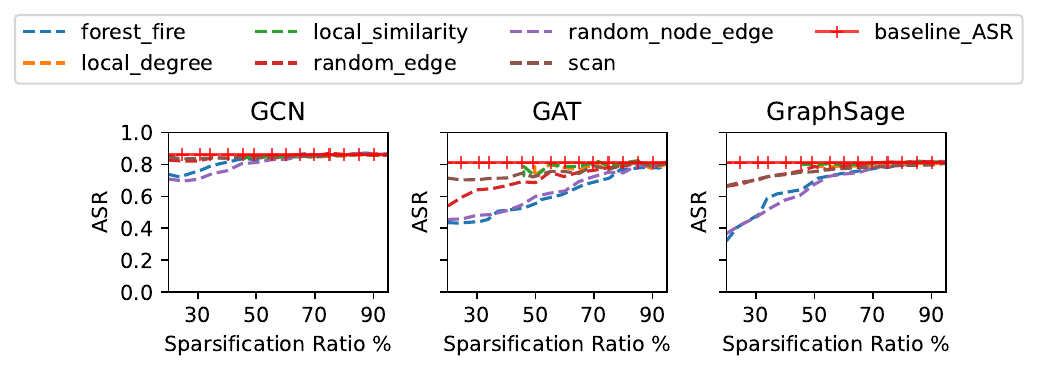}
    \caption{ASR on Pubmed}
    \label{fig:Sparsification Pubmed}
    \end{subfigure}
    \begin{subfigure}[b]{0.5\textwidth}
    \centering
    \includegraphics[width=\textwidth]{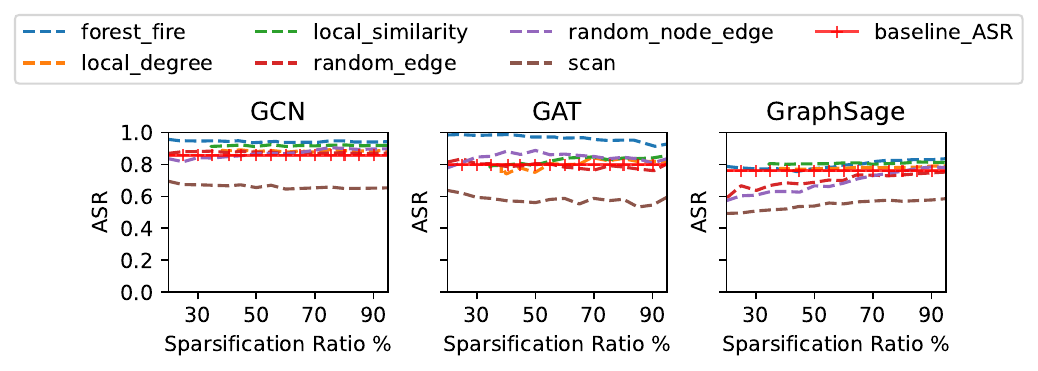}
    \caption{ASR on DBLP}
    \label{fig:Sparsification DBLP}
    \end{subfigure}
    \begin{subfigure}[b]{0.5\textwidth}
    \centering
    \includegraphics[width=\textwidth]{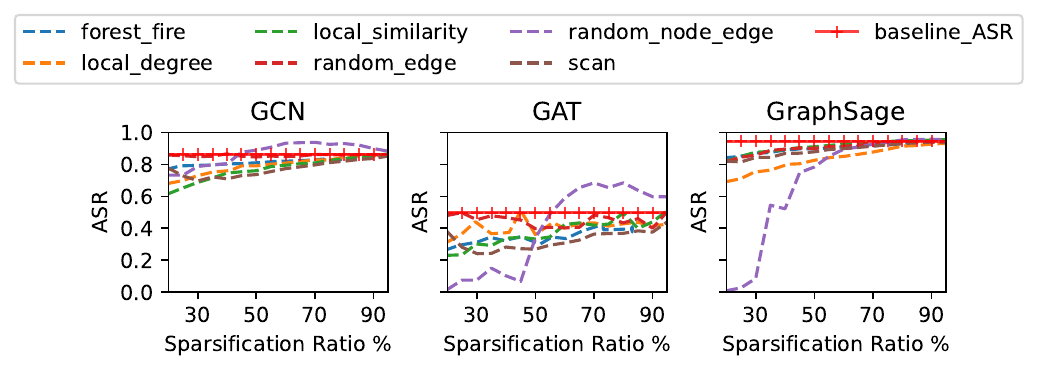}
    \caption{ASR on OGB-arxiv.}
    \label{fig:Sparsification OGB}
    \end{subfigure}
    \vspace{-10pt}
    \caption{ASR Under Different Graph Sparsification Methods.}
    \label{fig:asr_vs_sparsification}
    \vspace{-5pt}
\end{figure}

Table \ref{tab:sparsification_comparision} presents the ASR and ACC results using RNE for sparsification on different datasets under different backdoor attacks. Compared with the coarsening result given in Table \ref{tab:coarsening_comparision}, the ASR mitigation is less significant for sparsification when using sparsification ratios that maintain a drop in ACC within 2\%. On datasets such as Cora, Pubmed, sparsification has demonstrated better mitigation effects than Prune and Prune+LD methods. On the Cora dataset, the ASR of UGBA decreases from 93.38\% to 75.1\% at \( s = 0.3 \), and on the Pubmed dataset, the ASR for UGBA decreases from 83.07\% to 57.89\% at \( s = 0.3 \). In contrast, we observe a significant increase of ASR for the GTA attack under graph sparsification for Cora, Pubmed even within only 2\% ACC drop. We further examine the separate results for different models under UGBA attack and show the results in Figure \ref{fig:asr_vs_sparsification}. On the PubMed dataset, the ASR typically decreases as the sparsification ratio decreases across all three GNN models. However, on DBLP, we can also see that ASR increases compared to the baseline ASR. For example, using the Forest Fire method of sparsification on the GAT model results in an increase in ASR by more than 15\%. On the larger dataset OGB-arxiv, as shown in Figure \ref{fig:Sparsification OGB}, the ASR also increases by at least 10\% on the GAT model with RNE sparsification method. This highlights that \textbf{sparsification, under specific configurations, may inadvertently enhance the effectiveness of backdoor attacks}. Therefore, it is crucial to carefully evaluate and apply sparsification-based graph reduction when designing security-aware scalable GNN training systems.


\subsection{Trigger Analysis}
\label{Sec:triggeranalysis}
In this section, we present the result of the trigger analysis for graph coarsening and sparsification, respectively.
\subsubsection{Graph Coarsening}



\begin{table}
\centering
\setlength{\tabcolsep}{4pt}
\scriptsize
\caption{Merge Ratio ($m$\%), Label Change Ratio ($l$\%) and Feature Distance ($d$) of UGBA Triggers under Different Coarsening Ratios. `Null' indicates no coarsening.}
\begin{tabular}{l|ccc|ccc|ccc|ccc} 
\toprule
      & \multicolumn{3}{c|}{Cora} & \multicolumn{3}{c|}{Pubmed} & \multicolumn{3}{c}{DBLP}  & \multicolumn{3}{c}{OGB-arxiv}\\
      & $m$     & $l$     & $d$         & $m$     & $l$    & $d$            & $m$     & $l$     & $d$    & $m$     & $l$     & $d$     \\ 
\hline
$c$=0.3 & 63.9 & 21.5 & 1.5 & 86.6 & 2.9 & 0.8 & 80.4 & 14.5 & 0.6 & 86.3 & 16.5 & 32.5\\
$c$=0.5 & 39.8 & 15.0 & 1.4 & 85.9 & 2.8 & 0.8 & 74.2 & 0.7 & 0.4 & 34.2 & 17.5 & 59.0\\
$c$=0.7 & 18.5 & 10.3 & 1.8 & 84.1 & 2.3 & 0.8 & 66.4 & 0.6 & 0.5 & 2.5 & 15.8 & 79.0\\
$c$=0.9 & 0.0 & 1.9 & 2.0 & 71.3 & 1.3 & 1.0 & 33.2 & 0.1 & 0.7 & 25.7 & 2.5 & 63.4\\
\text{Null} & 0.0 & 0.0 & 2.4 & 0.0 & 0.0 & 1.5 & 0.0 & 0.0 & 2.0 & 0.0 & 0.0 & 44.2\\
\bottomrule
\end{tabular}
\label{Tab:trigger_analysis}
\end{table}

We perform the trigger analysis to understand how graph coarsening affects triggers injected by backdoor attacks. Table \ref{Tab:trigger_analysis} presents the results of three metrics defined in Section \ref{ssec:evaluationframework}. The merge ratio ($m$) indicates a substantial proportion of triggers being merged into super nodes, which could dilute the intended effect of the triggers. For example, on Pubmed, a coarsening ratio of 0.3 results in a merge ratio of 86.93\%, indicating significant merging of triggers. The label change ratio ($l$) demonstrates a significant reversion of poisoned labels to their original clean labels, highlighting the robustness of graph coarsening in mitigating label corruption. On the Cora dataset, a coarsening ratio of 0.3 results in a label change ratio of 23.58\%, indicating effective label restoration. Lastly, the feature distance ($d$) decreases as the coarsening ratio decreases, suggesting that the features of the triggers are aggregated with the attached nodes through coarsening. For instance, compared to non-coarsened graph, on the DBLP dataset, a coarsening ratio of 0.5 results in a feature distance decrease from 2.01 to 0.94, indicating a significant alteration of the trigger's features. Our analysis explains the efficacy of graph coarsening in mitigating poisoning attacks by disrupting the structural and feature characteristics of triggers. Table \ref{Tab. UGBA Trigger Analysis Different Rho} further shows trigger analysis results under varying poisoning ratios on Pubmed. It reveals that a slightly higher trigger merging ratio may be responsible for the decrease in ASR observed in Section \ref{Sec: trigger and size} as the poisoning ratio increases.

\begin{table}
\centering
\scriptsize
\caption{Trigger Analysis of UGBA under Different Poisoning Ratio, on Pubmed with Coarsening Ratio $c=70\%$.}
\begin{tabular}{l|rrr}
\toprule
       & \multicolumn{1}{l}{$m$(\%)} & \multicolumn{1}{l}{$l$(\%)} & \multicolumn{1}{l}{$d$}  \\
       \toprule
$\rho$=5\%  & 84.14                     & 2.3                       & 0.8                    \\
$\rho$=10\% & 85.86                     & 1.72                      & 0.94                   \\
$\rho$=15\% & 85.8                      & 1.19                      & 0.84   \\
\bottomrule
\end{tabular}
\label{Tab. UGBA Trigger Analysis Different Rho}
\vspace{-10pt}
\end{table}

\begin{table}
\centering
\scriptsize
\setlength{\tabcolsep}{4pt}
\caption{Trigger Analysis of GTA under Coarsening.}
\begin{tabular}{l|ccc|ccc|ccc|ccc} 
\toprule
      & \multicolumn{3}{c|}{Cora} & \multicolumn{3}{c|}{Pubmed} & \multicolumn{3}{c}{DBLP}  & \multicolumn{3}{c}{OGB-arxiv}\\
      & $m$     & $l$     & $d$         & $m$     & $l$    & $d$            & $m$     & $l$     & $d$    & $m$     & $l$     & $d$     \\ 
\hline
c=0.3                                & 67.6                                    & 19.6                                            & 23.7                                & 86.6                                    & 2.9                                             & 38.4                                & 80.4                                    & 14.5                                            & 106.5     &86.3&16.5&2148.9                          \\
c=0.5                                & 45.4                                    & 13.1                                            & 33.7                                & 85.9                                    & 2.8                                             & 38.8                                & 74.2                                    & 0.7                                             & 98.5      &34.2&17.5&3852.7                          \\
c=0.7                                & 20.4                                    & 10.3                                            & 35.0                                & 84.1                                    & 2.3                                             & 40.0                                & 66.4                                    & 0.6                                             & 110.7     &2.5&15.7&4950.3                          \\
c=0.9                                & 0.0                                     & 1.9                                             & 37.2                                & 71.3                                    & 1.3                                             & 48.7                                & 33.2                                    & 0.1                                             & 160.1     &25.7&2.5&4099.8                          \\
Null                                 & 0.0                                     & 0.0                                             & 30.7                                & 0.0                                     & 0.0                                             & 14.8                                & 0.0                                     & 0.0                                             & 7.6       &0&0&5031.2                           \\
\bottomrule
\end{tabular}
\label{Tab. GTA Trigger Analysis}
\end{table}

\begin{table}
\centering
\setlength{\tabcolsep}{4pt}
\scriptsize
\caption{Trigger Analysis of SBA-Gen under Coarsening.}
\begin{tabular}{l|ccc|ccc|ccc|ccc} 
\toprule
      & \multicolumn{3}{c|}{Cora} & \multicolumn{3}{c|}{Pubmed} & \multicolumn{3}{c}{DBLP}  & \multicolumn{3}{c}{OGB-arxiv}\\
      & $m$     & $l$     & $d$         & $m$     & $l$    & $d$            & $m$     & $l$     & $d$    & $m$     & $l$     & $d$     \\ 
\hline
c=0.3                                & 36.1                                    & 38.5                                            & 0.3                                  & 0.0                                     & 43.0                                             & 0.3                                  & 41.8                                    & 17.7                                            & 0.5              &0.7&43.4&4.9                     \\
c=0.5                                & 0.0                                     & 23.8                                            & 0.3                                  & 0.0                                     & 28.1                                             & 0.3                                  & 0.0                                     & 4.7                                             & 0.6              &0&26.2&3.53                     \\
c=0.7                                & 0.0                                     & 7.5                                             & 0.3                                  & 0.0                                     & 8.4                                              & 0.2                                  & 0.0                                     & 0.4                                             & 0.6              &0&9.4&2.1                     \\
c=0.9                                & 0.0                                     & 0.9                                             & 0.3                                  & 0.0                                     & 0.4                                              & 0.2                                  & 0.0                                     & 0.1                                             & 0.5              &0.0&2.3&1.7                     \\
Null                                 & 0.0                                     & 0.0                                             & 0.3                                  & 0.0                                     & 0.0                                              & 0.2                                  & 0.0                                     & 0.0                                             & 0.5              &0&0&1.4                     \\
\bottomrule
\end{tabular}
\label{Tab. SBA-Gen Trigger Analysis}
\end{table}

\begin{table}
\centering
\setlength{\tabcolsep}{4pt}
\scriptsize
\caption{Trigger Analysis of SBA-Samp under Coarsening.}
\begin{tabular}{l|ccc|ccc|ccc|ccc} 
\toprule
      & \multicolumn{3}{c|}{Cora} & \multicolumn{3}{c|}{Pubmed} & \multicolumn{3}{c}{DBLP}  & \multicolumn{3}{c}{OGB-arxiv}\\
      & $m$     & $l$     & $d$         & $m$     & $l$    & $d$            & $m$     & $l$     & $d$    & $m$     & $l$     & $d$     \\ 
\hline
c=0.3                                & 36.1                                    & 38.5                                            & 0.3                                  & 0.0                                     & 43.0                                             & 0.3                                  & 41.8                                    & 17.7                                            & 0.5               &0.7&43.4&4.82                    \\
c=0.5                                & 0.0                                     & 23.8                                            & 0.3                                  & 0.0                                     & 28.1                                             & 0.3                                  & 0.0                                     & 4.7                                             & 0.6               &0.0&26.2&3.5                    \\
c=0.7                                & 0.0                                     & 7.5                                             & 0.3                                  & 0.0                                     & 8.4                                              & 0.2                                  & 0.0                                     & 0.4                                             & 0.6               &0.0&9.4&2.1                    \\
c=0.9                                & 0.0                                     & 0.9                                             & 0.3                                  & 0.0                                     & 0.4                                              & 0.2                                  & 0.0                                     & 0.1                                             & 0.6               &0.0&2.3&1.6                    \\
Null                                 & 0.0                                     & 0.0                                             & 0.3                                  & 0.0                                     & 0.0                                              & 0.2                                  & 0.0                                     & 0.0                                             & 0.6               &0.0&0.0&1.4                    \\
\bottomrule
\end{tabular}
\label{Tab. SBA-Samp Trigger Analysis}
\end{table}

The results of trigger analysis for GTA and SBA attacks can be found in Table \ref{Tab. GTA Trigger Analysis},\ref{Tab. SBA-Gen Trigger Analysis},\ref{Tab. SBA-Samp Trigger Analysis}. Comparing these tables, we have the following observations. First, for GTA and UGBA, as the coarsening ratio decreases, the merge ratio increases. This explains why the ASR decreases as the coarsening ratio decreases. For the SBA-Gen and SBA-Samp triggers, the merge ratio often becomes zero at higher coarsening ratios, meaning that triggers are not merged into fewer super-nodes. This explains why coarsening often does not work on SBA triggers. Second, when the coarsening ratio is high (i.e., $c=$0.7 or 0.9), the label change ratio remains relatively low. This suggests that coarsening does not significantly clean the poisoned labels given high coarsening ratios. Third, the feature distance decreases as the coarsening ratio decreases in most cases. However, the feature distance for SBA triggers remains relatively stable, suggesting that SBA triggers are more resilient to feature space disruptions caused by coarsening.

\subsubsection{Graph Sparsification}
Our results in Section \ref{ssec:robustsparsification} show that graph sparsification is less effective in mitigating attacks than coarsening and the ASR does not necessarily decrease as the sparsification ratio decreases.
The trigger analysis results for UGBA, presented in Table. \ref{Tab. Sparsification Trigger Analysis}, provides further insight into this observation. While the prune ratio increases as the sparsification ratio decreases, the post-sparsification poisoning ratio remains very close to the initial poisoning ratio (default 5\%) and, in many cases, even increases slightly after sparsification. Table \ref{Tab. Sparsification Trigger Analysis GTA}, \ref{Tab. Sparsification Trigger Analysis SBA-Samp} and \ref{Tab. Sparsification Trigger Analysis SBA-Gen} present the trigger analysis results of GTA and SBA, which demonstrate similar trends to those observed with UGBA. For GTA, it is observed that the post-sparsification poisoning ratio on the Cora dataset is higher than the default poisoning ratio. Particularly at $s=0.9$, with a low prune ratio and a higher post-sparsification poisoning ratio, the ASR increased from 84.55\% to 99.90\%. This trend is consistent across other datasets, and also applies to the SBA attack. In addition to the observed increase in the poisoning ratio, graph sparsification maintains the poisoned labels and features unchanged and may significantly reduce node degrees. These factors combined elucidate why sparsification might inadvertently increase ASR, highlighting the need for careful consideration when implementing sparsification in security-sensitive contexts.

\begin{table}
\centering
\setlength{\tabcolsep}{3pt}
\scriptsize
\caption{Prune Ratio ($prune$ \%), Post-Sparsification Poisoning Ratio ($spar\_\rho$ \%) of UGBA Triggers under Different Sparsification Ratios.}
\vspace{-5pt}
\begin{tabular}{l|rr|rr|rr|rr} 
\toprule
      & \multicolumn{2}{c|}{Cora}                      & \multicolumn{2}{c|}{Pubmed}                    & \multicolumn{2}{c|}{DBLP}                       & \multicolumn{2}{c}{OGB-arxiv}                    \\
      & \multicolumn{1}{l}{$prune$} & \multicolumn{1}{l|}{$spar\_\rho$} & \multicolumn{1}{l}{$prune$} & \multicolumn{1}{l|}{$spar\_\rho$} & \multicolumn{1}{l}{$prune$} & \multicolumn{1}{l|}{$spar\_\rho$}  & \multicolumn{1}{l}{$prune$} & \multicolumn{1}{l}{$spar\_\rho$} \\ 
\hline
$s$=0.3 & 86.1 & 5.2 & 96.5 & 2.7 & 98.0 & 1.8 & 99.6 & 5.7 \\
$s$=0.5 & 50.0 & 5.7 & 87.1 & 4.4 & 88.1 & 3.5 & 96.4 & 6.0 \\
$s$=0.7 & 35.2 & 5.6 & 59.4 & 5.3 & 67.5 & 4.7 & 83.0 & 5.8 \\
$s$=0.9 & 11.1 & 5.5 & 24.0 & 5.7 & 26.7 & 5.4 & 42.3 & 5.4  \\
\bottomrule
\end{tabular}
\vspace{-5pt}
\label{Tab. Sparsification Trigger Analysis}
\end{table}

\begin{table}
\centering
\setlength{\tabcolsep}{3pt}
\scriptsize
\caption{Trigger Analysis of GTA under Sparsification.}
\vspace{-5pt}
\begin{tabular}{l|rr|rr|rr|rr} 
\toprule
      & \multicolumn{2}{c|}{Cora}                      & \multicolumn{2}{c|}{Pubmed}                    & \multicolumn{2}{c|}{DBLP}                       & \multicolumn{2}{c}{OGB-arxiv}                    \\
      & \multicolumn{1}{l}{$prune$} & \multicolumn{1}{l|}{$spar\_\rho$} & \multicolumn{1}{l}{$prune$} & \multicolumn{1}{l|}{$spar\_\rho$} & \multicolumn{1}{l}{$prune$} & \multicolumn{1}{l|}{$spar\_\rho$}  & \multicolumn{1}{l}{$prune$} & \multicolumn{1}{l}{$spar\_\rho$} \\ 
\hline
$s$=0.3 & 63.6 & 5.3 & 80.4 & 3.1 & 76.7 & 2.3 & 99.6 & 5.8 \\
$s$=0.5 & 39.7 & 5.2 & 67.1 & 4.3 & 54.7 & 3.8 & 96.0 & 5.9 \\
$s$=0.7 & 31.9 & 5.5 & 44.4 & 5.3 & 33.8 & 5.2 & 82.9 & 5.8 \\
$s$=0.9 & 12.4 & 5.4 & 20.4 & 5.8 & 11.7 & 5.5 & 41.6 & 5.4  \\
\bottomrule
\end{tabular}
\vspace{-5pt}
\label{Tab. Sparsification Trigger Analysis GTA}
\end{table}

\begin{table}
\centering
\setlength{\tabcolsep}{3pt}
\scriptsize
\caption{Trigger Analysis of SBA-Samp under Sparsification.}
\vspace{-5pt}
\begin{tabular}{l|rr|rr|rr|rr} 
\toprule
      & \multicolumn{2}{c|}{Cora}                      & \multicolumn{2}{c|}{Pubmed}                    & \multicolumn{2}{c|}{DBLP}                       & \multicolumn{2}{c}{OGB-arxiv}                    \\
      & \multicolumn{1}{l}{$prune$} & \multicolumn{1}{l|}{$spar\_\rho$} & \multicolumn{1}{l}{$prune$} & \multicolumn{1}{l|}{$spar\_\rho$} & \multicolumn{1}{l}{$prune$} & \multicolumn{1}{l|}{$spar\_\rho$}  & \multicolumn{1}{l}{$prune$} & \multicolumn{1}{l}{$spar\_\rho$} \\ 
\hline
$s$=0.3 & 57.9 & 5.8 & 74.5 & 2.7 & 77.9 & 2.4 & 99.2 & 5.7\\
$s$=0.5 & 42.3 & 5.4 & 55.0 & 4.8 & 50.6 & 4.0 & 91.4 & 6.1\\
$s$=0.7 & 25.0 & 5.7 & 36.9 & 5.7 & 32.7 & 5.3 & 66.4 & 5.9\\
$s$=0.9 & 7.4 & 5.5 & 14.3 & 5.7 &  12.7 & 5.5 & 22.3 & 5.5\\
\bottomrule
\end{tabular}
\vspace{-5pt}
\label{Tab. Sparsification Trigger Analysis SBA-Samp}
\end{table}

\begin{table}
\centering
\setlength{\tabcolsep}{3pt}
\scriptsize
\caption{Trigger Analysis of SBA-Gen under Sparsification.}
\vspace{-5pt}
\begin{tabular}{l|rr|rr|rr|rr} 
\toprule
      & \multicolumn{2}{c|}{Cora}                      & \multicolumn{2}{c|}{Pubmed}                    & \multicolumn{2}{c|}{DBLP}                       & \multicolumn{2}{c}{OGB-arxiv}                    \\
      & \multicolumn{1}{l}{$prune$} & \multicolumn{1}{l|}{$spar\_\rho$} & \multicolumn{1}{l}{$prune$} & \multicolumn{1}{l|}{$spar\_\rho$} & \multicolumn{1}{l}{$prune$} & \multicolumn{1}{l|}{$spar\_\rho$}  & \multicolumn{1}{l}{$prune$} & \multicolumn{1}{l}{$spar\_\rho$} \\ 
\hline
$s$=0.3 & 57.9 & 5.5 & 75.8 & 3.2 & 71.7 & 2.2 & 99.2 & 5.8\\
$s$=0.5 & 45.5 & 5.4 & 51.7 & 4.5 & 50.3 & 4.2 & 92.1 & 6.0\\
$s$=0.7 & 33.3 & 5.7 & 40.2 & 5.6 & 37.1 & 5.4 & 65.9 & 5.9\\
$s$=0.9 & 6.6  & 5.4 & 13.4 & 5.8 & 12.9 & 5.5 & 22.7 & 5.5\\
\bottomrule
\end{tabular}
\vspace{-5pt}
\label{Tab. Sparsification Trigger Analysis SBA-Gen}
\end{table}

\subsection{Poisoned Node Analysis}
\label{Sec: poisoned nodes}
\begin{figure}
    \centering
    \begin{subfigure}[b]{0.5\textwidth}
    \centering
    \includegraphics[width=\textwidth]{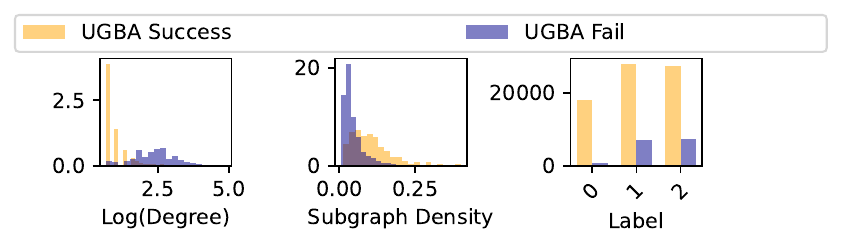}
    \caption{UGBA attack under no graph reduction}
    \label{Fig: Node_Analysis_Pubmed_UGBA}
    \end{subfigure}
    \begin{subfigure}[b]{0.5\textwidth}
    \centering
    \includegraphics[width=\textwidth]{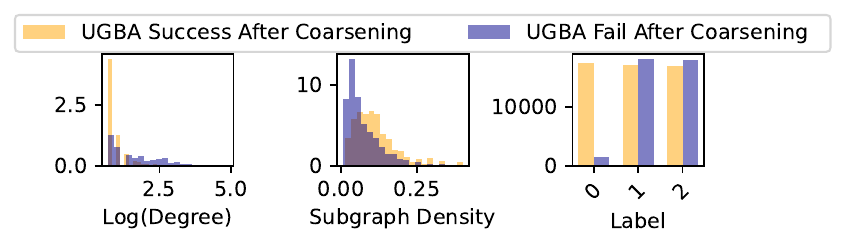}
    \caption{UGBA attack under graph coarsening}
    \label{Fig: Node_Analysis_Pubmed_UGBA_Coarsening}
    \end{subfigure}
    \begin{subfigure}[b]{0.5\textwidth}
    \centering
    \includegraphics[width=\textwidth]{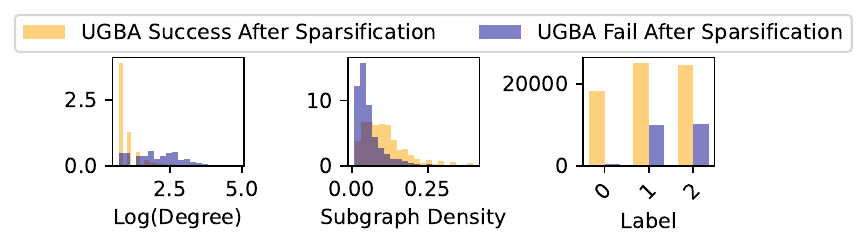}
    \caption{UGBA attack under graph sparsification}
    \label{Fig: Node_Analysis_Pubmed_UGBA_Sparsification}
    \end{subfigure}
    \caption{Analysis of Poisoned Node Distributions.}
    \vspace{-10pt}
\end{figure}

To understand how graph reduction affects poisoned nodes differently, we collected a range of features for each targeted node. These features included the node's degree, the density of its 2-hop subgraph, its ground truth label, and the target GNN model (GCN, GAT, and GraphSAGE). We show our experiments on the Pubmed dataset, which contains three labels (label 0, label 1, and label 2). We use label 0 as the target label in UGBA. Figure \ref{Fig: Node_Analysis_Pubmed_UGBA} shows the distribution of these features for both the successfully attacked nodes and failed nodes using UGBA in the case of no graph reduction. The log(degree) and subgraph density are shown as normalized distributions.
Figure \ref{Fig: Node_Analysis_Pubmed_UGBA_Coarsening} and \ref{Fig: Node_Analysis_Pubmed_UGBA_Sparsification} depict the distribution of these features after coarsening and sparsification, respectively. 

Comparing Figure \ref{Fig: Node_Analysis_Pubmed_UGBA_Coarsening} and \ref{Fig: Node_Analysis_Pubmed_UGBA_Sparsification} with Figure \ref{Fig: Node_Analysis_Pubmed_UGBA},
we can see that graph coarsening effectively mitigates backdoor attacks on target nodes with lower degrees, whereas sparsification yields a distribution similar to that of the original UGBA (i.e., under no graph reduction). This indicates that graph coarsening is more effective at safeguarding lower-degree nodes than sparsification. Additionally, the label distribution under graph coarsening reveals that this method uniformly protects nodes with ground truth labels of both 1 and 2, underscoring its consistent effectiveness across different node classes.
Furthermore, the similarity of the distributions with sparsification to those of the original UGBA indicates that sparsification does not significantly alter the attack’s distribution patterns.

\section{Conclusion and Future work}
Our experiments with multiple datasets and attack scenarios demonstrate the effect of graph reduction in GNN robustness against existing backdoor attacks. Graph reduction tends to be more effective in reducing ASRs when paired with GAT and GraphSage architectures compared to GCN, or when implemented with a lower reduction ratio. Graph coarsening consistently mitigates backdoor attacks effectively, although the efficacy can vary among different coarsening methods. Specifically, graph coarsening is particularly effective at protecting the most vulnerable nodes within a network, notably low-degree nodes, whereas sparsification does not provide similar protection. In fact, graph sparsification may exacerbate vulnerabilities, as evidenced by increased ASRs that surpass those in systems without any sparsification. This highlights a critical concern: while sparsification may offer computational benefits, it can also increase security risks, underscoring the need for careful consideration of graph reduction strategies in the development of secure GNN systems.

In our future work, we aim to evaluate a broader range of attack types, including both evasion and poisoning attacks, within graph reduction systems. We plan to explore the development of new graph reduction algorithms that integrate robustness properties with the goal of enhancing the mitigation effects against backdoor attacks. This will include a particular focus on addressing and circumventing the potential risks associated with sparsification, ensuring that any reductions do not inadvertently increase the system’s vulnerability to attacks. It is also interesting to see if there exist more powerful attacks capable of overcoming the mitigation effects of graph reduction.

\bibliographystyle{ACM-Reference-Format}
\bibliography{ref}

\appendix 
\section{Appendix}

\subsection{Descriptions of Graph Reduction Methods}
\label{app:methods}
In this paper, NetworKit\cite{NetworkKit}, an open-source python toolkit is used for the implementation of different reduction methods. 

\noindent\textbf{Graph Coarsening:}~~
Six methods for graph coarsening are listed as follows:
\begin{itemize}[topsep=1pt, leftmargin=*]
    \item Variation Neighbourhoods (\textit{VN}), Variation Cliques (\textit{VC}), Variation Edges (\textit{VE}) \cite{loukas2019graph}: These methods are predicated on spectral principles. The process begins with calculating the graph Laplacian matrix $\mathbf{L}$. Given a target graph dimension $n$, the objective is to derive a coarsened Laplacian matrix $\mathbf{L}_c$ of dimension $n \times n$ that approximates $\mathbf{L}$ with graph information loss $\epsilon$ below a predetermined threshold. The distinction among the \textit{VN}, \textit{VC}, and \textit{VE} methodologies lies in how they select node sets for contraction. For the \textit{VN} strategy, each vertex along with its neighbors forms a candidate set; for \textit{VC}, all maximal cliques identified using the Bron-Kerbosch algorithm are considered as separate candidate sets; and for \textit{VE}, individual edges serve as candidate sets. After sorting these sets based on cost, recursive computations are performed. First, the candidate set with the lowest cost is selected, and then the vertices within these sets are coarsened. The recursion stops when the updated $\mathbf{L}_c$ reaches the desired size.
    \item Heavy Edge Matching: For the \textit{Heavy Edge Matching} approach, edge pairs are selected for contraction at each coarsening level by computing the Maximum Weight Matching, wherein the weight of an edge pair is determined based on the maximum vertex degree within the pair \cite{loukas2018spectrally}. This strategy tends to contract edges peripheral to the main graph body, thereby preserving the core structural integrity of the graph.
    \item Algebraic JC:  It calculates algebraic distances as weights based on each candidate set of edges, where the distances are computed from test vectors, each of which is computed out of scans of the Jacobi relaxation \cite{ron2011relaxation}.
    \item Kron Reduction \cite{dorfler2012kron}: At each coarsening stage of the \textit{Kron Reduction} method, a subset of vertices is selected, identified by the positive entries of the Laplacian matrix's final eigenvector. Through Kron Reduction, the graph's size is reduced, targeting the preservation of its spectral characteristics for efficient analysis of extensive networks.
\end{itemize}

\noindent\textbf{Graph Sparsification:}~~
\begin{itemize}[topsep=1pt,leftmargin=*]
\item Random Edge (RE): It simply selects edges to keep in the sparsified graph uniformly at random.
\item Random Node Edge (RNE): It uniformly selects both edges and nodes to keep in the sparsified graph at random.
\item Local Degree~\cite{hamann2016structure}: It ranks the neighboring nodes by their degree and selects a fraction of the top-ranked neighbors while making sure that each node retains at least one edge.
\item Local Similarity~\cite{satuluri2011local}: It calculates the Jaccard similarity scores between vertex and its neighbors, and select edges with the highest similarity scores locally for inclusion in the sparsified graph. 
\item Forest Fire~\cite{leskovec2007graph}: It firstly chooses a random seed node, then iteratively adding neighbor nodes with the edge between them until every node were selected.
\item Scan\cite{xu2007scan}: It sorts the edges by calculating the SCAN similarity score for all pairs of vertices in the graph and selects the edges with the highest similarity scores for inclusion in the sparsified graph.
\end{itemize}

\subsection{Supplementary Tables}

\begin{table}[H]
\scriptsize
\centering
\caption{Node Classification Dataset Description}
\begin{tabular}{lcccc}
\hline
Datasets   & \#Nodes & \#Edges  & \#Feature & \#Classes \\ \hline
Cora       & 2,708   & 5,429    & 1,443     & 7         \\
Pubmed     & 19,717  & 44,338   & 500       & 3         \\
DBLP     & 17,716  & 105,734  & 1,639       & 4         \\
OGB-arxiv  & 169,343 & 1,166,243 & 128      & 40        \\ \hline
\end{tabular}
\label{table:dataset}
\end{table}

\begin{table}[H]
\centering
\caption{Memory Reduction with Coarsening for GraphSage}
\scriptsize
\begin{tabular}{cc|ccc} 
\toprule
      Dataset     & \multicolumn{1}{l|}{orig (MB)} & c=0.7         & c=0.5         & c=0.3          \\ 
\hline
Cora       & 184            & 160 ($\downarrow$12.8\%)   & 145 ($\downarrow$21.1\%)   & 125 ($\downarrow$31.9\%)    \\
Pubmed     & 373            & 320 ($\downarrow$14.3\%)   & 274 ($\downarrow$26.6\%)   & 233 ($\downarrow$37.6\%)   \\
DBLP       & 1132           & 941 ($\downarrow$26.9\%)   & 781 ($\downarrow$31.1\%)   & 663 ($\downarrow$41.5\%)   \\
Physics    & 19933          & 15539 ($\downarrow$22.2\%) & 12185 ($\downarrow$38.9\%) & 9554 ($\downarrow$52.1\%)  \\
OGB-arxiv & 1559           & 1229 ($\downarrow$21.2\%)  & 993 ($\downarrow$36.4\%)   & 759 ($\downarrow$51.4\%)   \\
\bottomrule
\end{tabular}
\label{Fig. Memory Coarsening}
\end{table}

\begin{table}[H]
\centering
\caption{Memory Reduction with Sparsification for GraphSage}
\scriptsize
\begin{tabular}{cc|ccc} 
\toprule
      Dataset     & \multicolumn{1}{l|}{orig (MB)} & s=0.7         & s=0.5         & s=0.3          \\ 
\hline
Cora       & 184             &   163 ($\downarrow$11.3\%)    &   148 ($\downarrow$19.5 \%)  &   131 ($\downarrow$28.7 \%)    \\
Pubmed     & 373             &   305 ($\downarrow$18.4\%)   &   270 ($\downarrow$27.7\%)  &   234 ($\downarrow$37.4\%)   \\
DBLP       & 1132            &   880 ($\downarrow$22.7\%)   &   743 ($\downarrow$34.4\%)  &   605 ($\downarrow$26.6\%)   \\
Physics    & 19933           &   14367 ($\downarrow$26.6\%) &   11428 ($\downarrow$42.7\%) &   8218 ($\downarrow$58.8\%)  \\
OGB-arxiv & 1559            &   1229 ($\downarrow$21.2\%)   &   993 ($\downarrow$36.3\%)  &   759 ($\downarrow$51.4\%)   \\
\bottomrule
\end{tabular}
\label{Fig. Memory Sparsification}
\end{table}

\end{document}